\documentclass[sigconf, 9pt, nonacm]{acmart}

\AtBeginDocument{%
  }

\usepackage{balance}
\usepackage{subfig}
\usepackage{xspace}
\usepackage{tabularx}
\usepackage{multirow}

\newcommand{\sys}{mmCounter\xspace}
\newcommand\rnote[1]{\textcolor{black}{#1}}
\newcommand{\parlabel}[1]{\vspace{0.5em}\noindent\textbf{#1}}

\newcolumntype{Y}{>{\centering\arraybackslash}X}

\begin{document}

\title{\sys: Static People Counting in Dense Indoor Scenarios using mmWave Radar}

\author{Tarik Reza Toha}
\email{ttoha12@cs.unc.edu}
\orcid{0000-0002-6529-3487}
\affiliation{%
  \institution{University of North Carolina}
  \city{Chapel Hill}
  \state{NC}
  \country{USA}
}

\author{Shao-Jung (Louie) Lu}
\email{louielu@cs.unc.edu}
\orcid{0009-0007-8530-9283}
\affiliation{%
  \institution{University of North Carolina}
  \city{Chapel Hill}
  \state{NC}
  \country{USA}
}

\author{Shahriar Nirjon}
\email{nirjon@cs.unc.edu}
\orcid{0000-0003-1443-1146}
\affiliation{%
  \institution{University of North Carolina}
  \city{Chapel Hill}
  \state{NC}
  \country{USA}
}

\renewcommand{\shortauthors}{Toha et al.}

\begin{abstract}
mmWave radars struggle to detect or count individuals in dense, static (non-moving) groups due to limitations in spatial resolution and reliance on movement for detection. We present \sys, which accurately counts static people in dense indoor spaces (up to three people per square meter). \sys achieves this by extracting ultra-low frequency ($<$ 1 Hz) signals, primarily from breathing and micro-scale body movements such as slight torso shifts, and applying novel signal processing techniques to differentiate these subtle signals from background noise and nearby static objects. Our problem differs significantly from existing studies on breathing rate estimation, which assume the number of people is known a priori. In contrast, \sys utilizes a novel multi-stage signal processing pipeline to extract relevant low-frequency sources along with their spatial information and map these sources to individual people, enabling accurate counting. Extensive evaluations in various environments demonstrate that \sys delivers an 87\% average F1 score and 0.6 mean absolute error in familiar environments, and a 60\% average F1 score and 1.1 mean absolute error in previously untested environments. It can count up to seven individuals in a three square meter space such that there is no side-by-side spacing and only a one-meter front-to-back distance.
\end{abstract}





\maketitle

\section{Introduction}

Accurate indoor people counting is vital for improving efficiency, safety, and user experience across various domains~\cite{li2023counting,huang2023application,ren2023grouped}. It drives smarter, more adaptive environments in diverse, high-impact applications. In intelligent buildings, it enables real-time optimization of energy use and resource allocation, supporting sustainability \cite{tang2020occupancy}. Retailers analyze customer flow to enhance service and sales \cite{myint2021people}. Public transportation hubs rely on it to manage crowd control and reduce congestion \cite{ren2023grouped}. In healthcare, it helps monitor patient movement and control infections \cite{li2023counting}. Event management benefits from precise attendee counts for logistical planning and safety compliance \cite{toha2022lc}. 


\rnote{People counting techniques employ various sensors, each with distinct advantages. Camera-based systems \cite{chen2022counting, shu2022crowd} use image processing and computer vision for high accuracy but pose privacy risks and struggle in low light. Thermal sensors \cite{xie2023efficient, hagenaars2020single} detect heat signatures, offering a privacy-preserving alternative, but their accuracy depends on room temperature \cite{bao2021cnn} and lighting \cite{liu2021cross}, making them unreliable in controlled spaces. In dense settings, overlapping heat signatures hinder individual distinction, challenging thermal-based people counting \cite{collini2024flexible, pan2023cginet}. In contrast, mmWave radar \cite{marco2024mmwave, ren2023grouped} excels in dense crowds and through some obstacles. It operates reliably in all lighting conditions, maintains high accuracy, and preserves privacy, making it ideal for complex indoor environments.}


\rnote{Several recent studies have leveraged mmWave radar for people counting in various scenarios \cite{ren2023grouped, li2023counting, ti2024demo}. However, these methods are limited to counting only moving individuals, as they rely on spatio-temporal features by tracking point clouds, which require subjects to be in motion. Their performance degrades in dense environments where closely positioned individuals ($>$2 people/m$^2$) cause overlapping radar reflections due to the radar's limited spatial resolution. Consequently, the point clouds of multiple closely spaced individuals merge, making individual tracking infeasible. To mitigate this issue, \cite{marco2024mmwave} introduced a deep-learning-based approach for counting individuals in dense outdoor settings. However, this method also shares the same limitation: it depends on movement and is unable to count stationary individuals.}


\rnote{Existing approaches \cite{li2023counting, marco2024mmwave} place mmWave radar at room entrances, similar to doorway sensors, and rely on continuous monitoring to track entries and exits. However, they must be positioned at entry points \cite{korany2021counting} and only provide an aggregate count of people inside, without identifying their spatial distribution. As a result, these methods are unsuitable for scenarios requiring real-time occupancy snapshots in static environments, such as waiting rooms or prayer halls, where accurate counts are essential for applications like HVAC adjustment and service optimization \cite{korany2021counting}. These limitations highlight the need for a mmWave radar system capable of accurately counting people in both static and dense indoor settings.}


In this paper, we propose a mmWave radar-based people counting system, named \sys, capable of counting up to three stationary individuals per square meter. Specifically, \sys can accurately count individuals standing with nearly zero side-to-side and one-meter front-to-back interpersonal spacing. While this interpersonal spacing aligns with established social studies of personal space~\cite{kroczek2020interpersonal}, it is not a limitation of our system, but rather a realistic upper bound based on human spatial behavior in dense settings. We address three key technical challenges in counting static individuals in such dense indoor scenarios:

\begin{figure*}[!t]
\centering
\subfloat[Spectrogram (static period)]
{\includegraphics[width=0.24\textwidth]{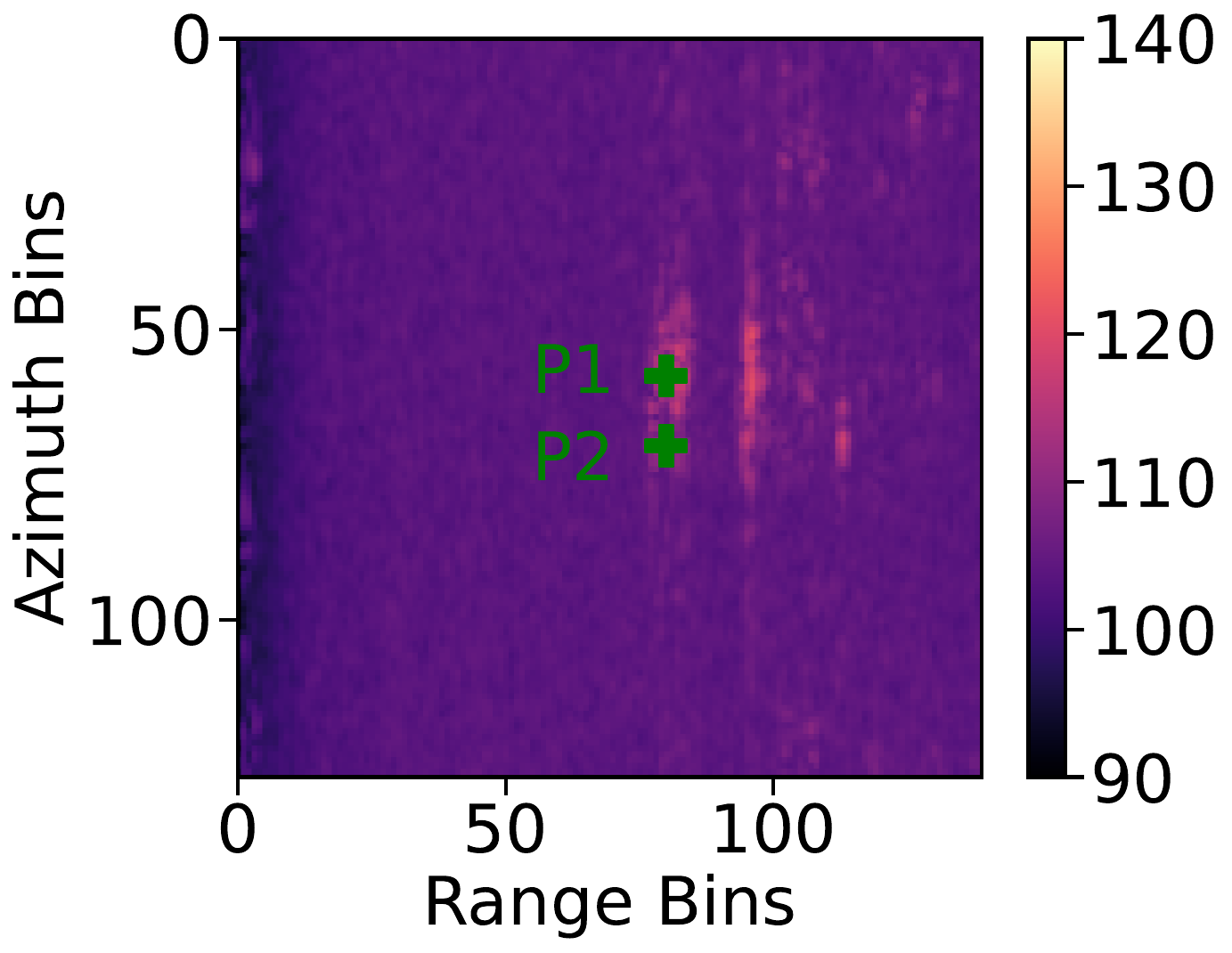}
\label{fig:azi0}}
\hfil
\subfloat[Spectrogram (moving period)]
{\includegraphics[width=0.24\textwidth]{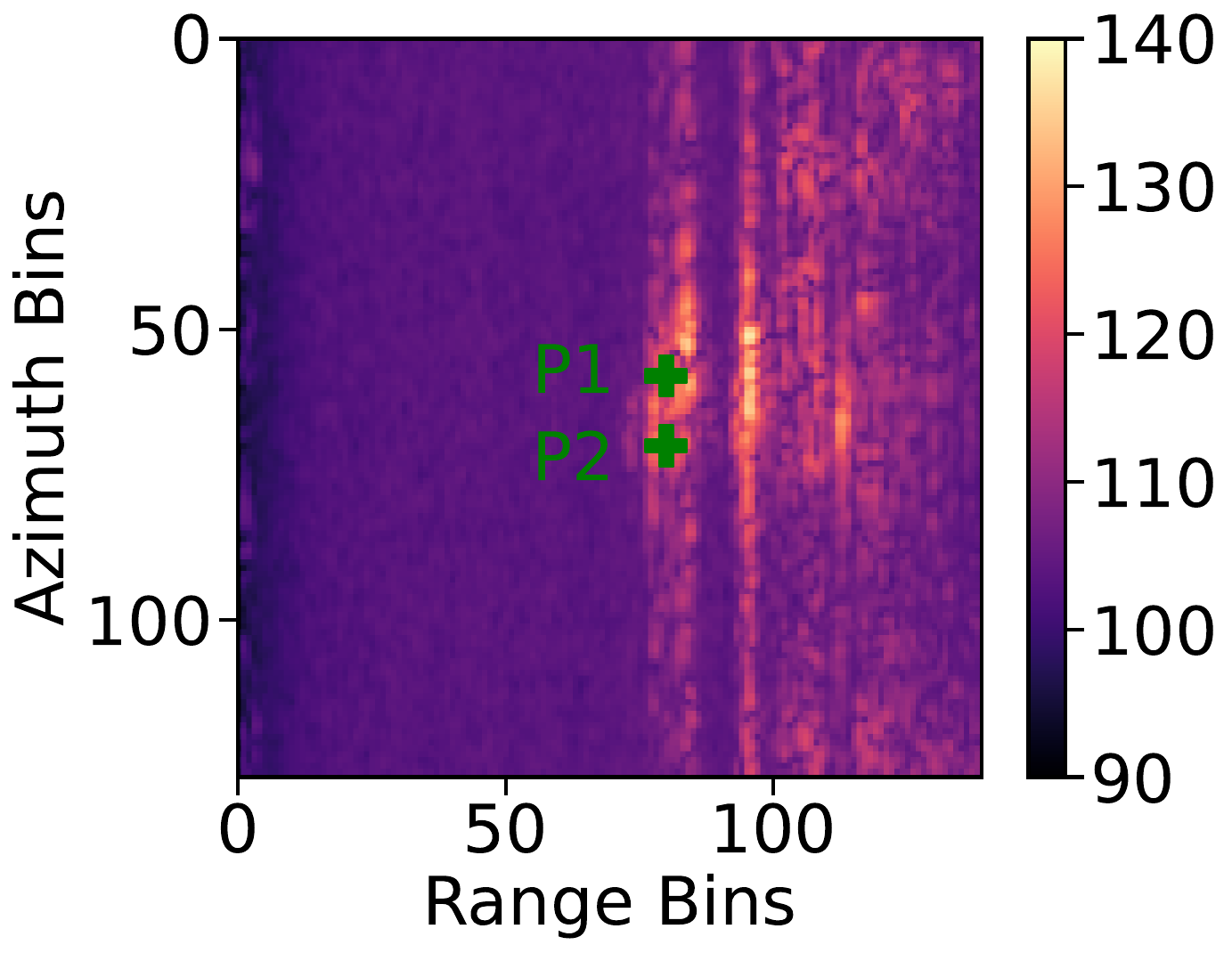}
\label{fig:azi1}}
\hfil
\subfloat[Point cloud (static period)]
{\includegraphics[width=0.24\textwidth]{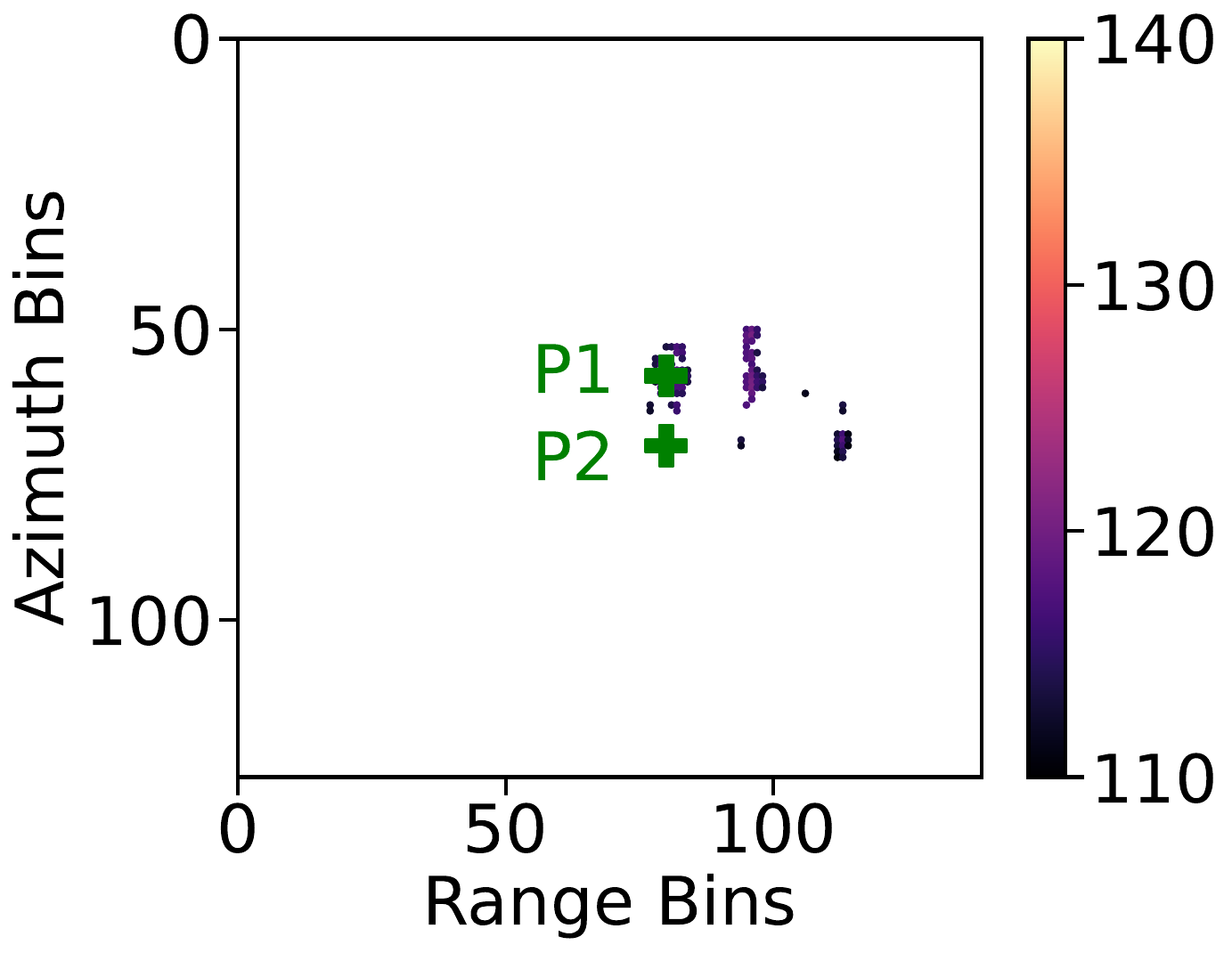}
\label{fig:pts0}}
\hfil
\subfloat[Point cloud (moving period)]
{\includegraphics[width=0.24\textwidth]{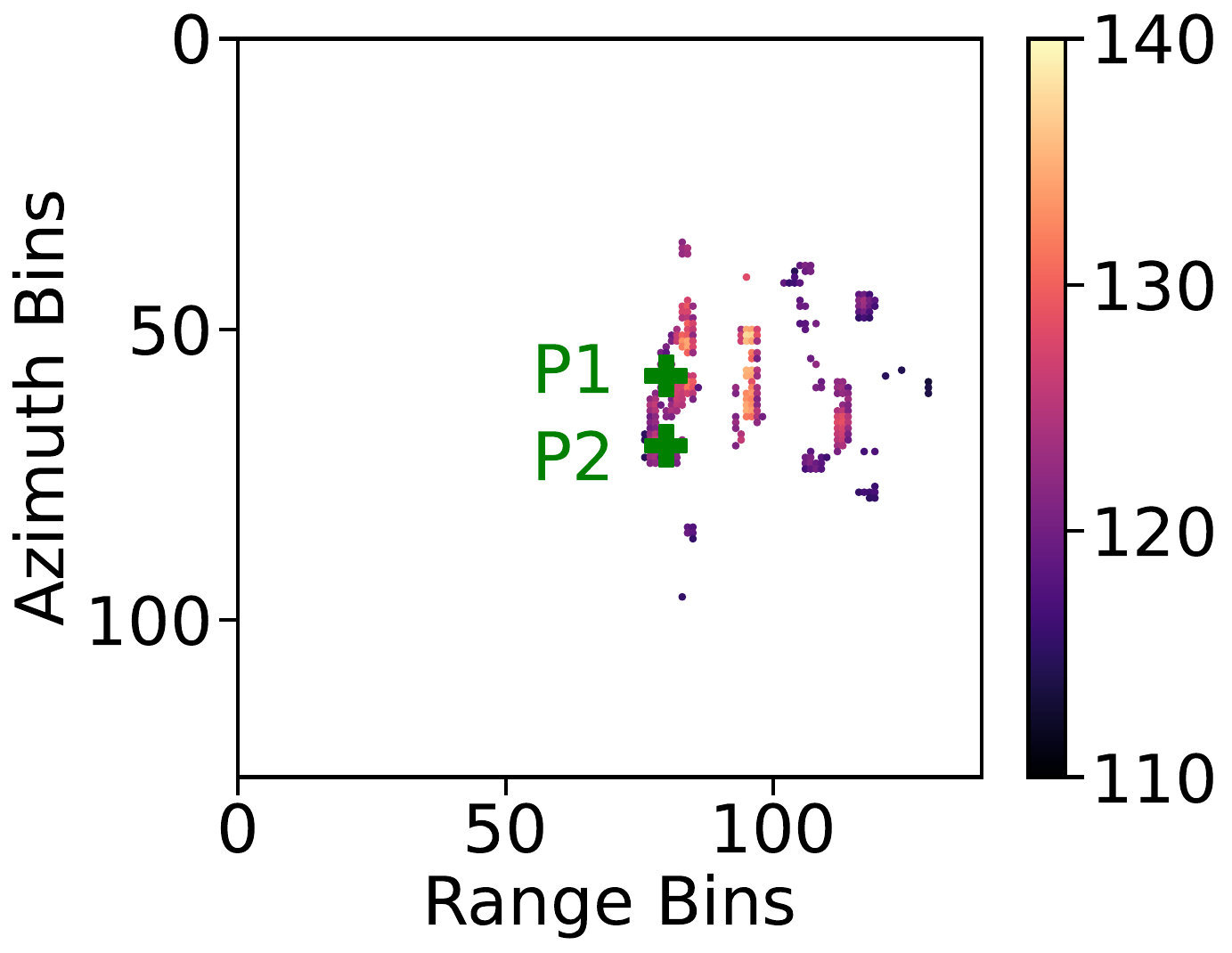}
\label{fig:pts1}}
\vspace{-3mm}
\caption{Spectrograms and point clouds of two individuals during a static period (standing) followed by a moving period}
\vspace{-3mm}
\end{figure*}

First, mmWave radar typically requires minimal movement to differentiate a person from the background. Ideally, a fully static individual cannot be detected. However, static (e.g., standing) individuals generate micro-motions due to physiological processes like breathing, heartbeat, and slight body swaying. These micro-motions contain critical information that can be used to count people. Existing studies employ the phase-change method to isolate breathing signals from individuals~\cite{gao2022real, wang2020remote}, but this method works only for a single person positioned close to the radar. The challenge here is to \emph{reliably detect micro-motion from multiple individuals at a distance.}

Second, micro-motion consists of a combination of periodic and random signals. Periodic signals, such as breathing and heartbeat, directly correlate with the number of individuals present. For accurate people counting, breathing signals are especially useful since breathing induces measurable vibrations in the torso (chest-belly) area. \cite{yue2018extracting} used Independent Component Analysis (ICA) \cite{hyvarinen2000independent} to separate micro-motion into breathing and non-breathing components, but they assume prior knowledge of the number of individuals~\cite{wang2020respiration}. The challenge is \emph{extracting breathing signals when the number of people is unknown.} 

Third, not all extracted breathing signals are unique, and there is no straightforward one-to-one mapping between breathing signals and the number of individuals. A single person can produce multiple breathing signals, resulting in a many-to-one relationship. The challenge here is to \emph{accurately map breathing signals to individual people.} This breathing-to-people relationship is non-trivial, and no existing study has effectively addressed this problem. 

We propose a series of novel techniques to address the three key challenges of static people counting in dense scenarios. Accordingly, the novelties of our system are stated below:
\begin{enumerate}
    \item We design a robust micro-displacement estimation algorithm to extract micro-motion signals from multiple individuals standing at least two meters away.

    \item We design a novel breathing source extraction algorithm to extract both high-quality breathing sources and their spatial distributions when the number of people is unknown.

    \item We develop a self-attention-based deep-learning architecture to determine the number of unique breathing groups (and thus the number of people) based on their spatial distributions.
\end{enumerate}


We comprehensively evaluate \sys in two indoor environments, varying the number of individuals, spatial configurations, and interpersonal distances. In a known environment, \sys achieves 88.1\% average precision, 85.6\% average recall, 86.8\% average F1 score, 0.6 mean absolute error, and 2.3 mean squared error. In an unseen, more challenging environment, \sys maintains 67.6\% average precision, 53.3\% average recall, 59.6\% average F1 score, 1.1 mean absolute error, and 2.3 mean squared error, demonstrating its generalizability despite environmental variations.  

\sys accurately counts up to seven individuals in an extreme corner case with nearly zero side-by-side spacing and a one-meter front-to-back distance within a three-square-meter area, pushing the system’s limits. Additionally, under various blocking scenarios, \sys achieves up to 86.5\% average precision, 70.2\% average recall, 77.5\% average F1 score, 0.4 mean absolute error, and 0.6 mean squared error, demonstrating its robustness.

\begin{figure*}[!t]
\centering
\includegraphics[width=\textwidth]{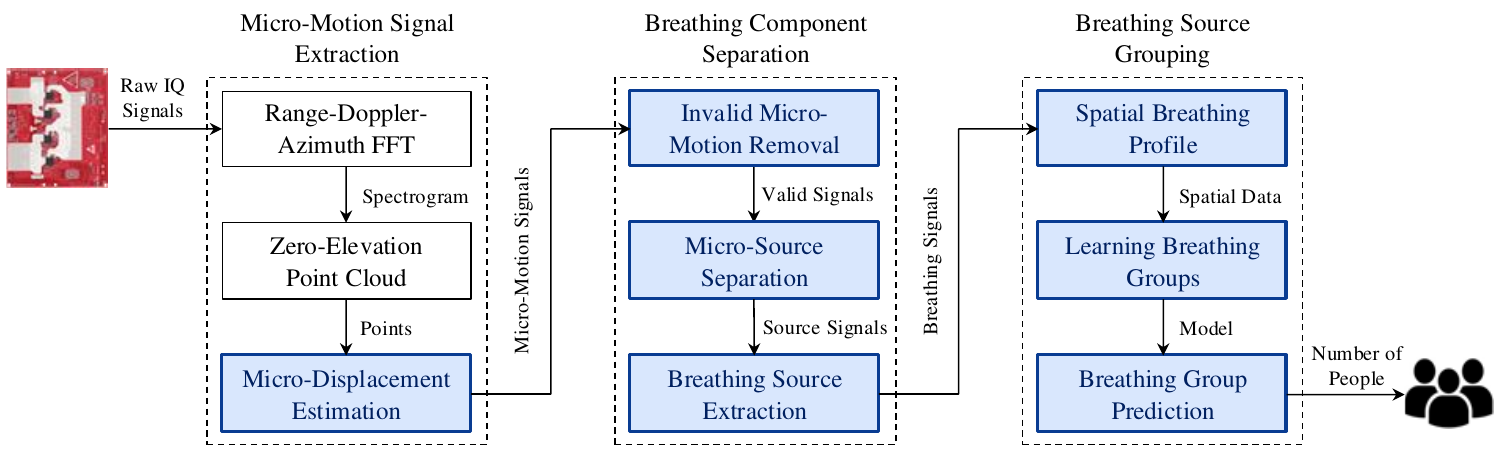}
\vspace{-3mm}
\caption{Block diagram of our proposed people counting system, where blue-colored blocks are newly proposed}
\label{fig:blocks}
\vspace{-3mm}
\end{figure*}

\section{Preliminary Study}
\label{sec:prelim}

We examine the challenge of counting static people compared to moving individuals and demonstrate that state-of-the-art models fail to address this issue.

\subsection{Static People in Radar Scene}
This experiment demonstrates why counting static individuals with mmWave radar is more challenging than counting moving ones. Minimal motion from static people limits the radar’s ability to generate point clouds, a primary data source in existing people-counting systems. By comparing static and moving periods, we assess the radar’s detection capabilities and its limitations in identifying non-moving individuals.

We recorded a scene where two individuals, P1 and P2, stood side-by-side with a 6-inch separation during the \emph{static} period, then walked together during the \emph{moving} period. Using FFT for frequency analysis and CFAR for thresholding, we generate frequency spectrograms and point clouds for both periods to evaluate the radar’s performance in static and dynamic contexts. The radar configuration parameters are listed in Table \ref{tab:param}.

During the static period, reflection intensity is significantly lower, and point clouds are sparse, with P2 producing no detectable points, as shown in Figures~\ref{fig:azi0} and~\ref{fig:pts0}. In contrast, during the moving period, both reflection intensity and point cloud density increase, as seen in Figures~\ref{fig:azi1} and~\ref{fig:pts1}. These results confirm that mmWave radar relies heavily on movement for detection, as static individuals generate insufficient reflections for reliable tracking.

Our results reveal a fundamental limitation of existing mmWave radar-based people-counting methods, which primarily track point clouds generated by moving individuals. Approaches like \cite{li2023counting} and \cite{marco2024mmwave} cluster and classify these point clouds, making them ineffective for static individuals. This experiment highlights the necessity for novel techniques that enable accurate counting in static scenarios without relying on movement.

\subsection{End-to-End Deep Learning}
Since static individuals do not generate sufficient points in the point cloud, we adopt an end-to-end deep learning approach inspired by \cite{marco2024mmwave} to directly classify frequency spectrograms for static people counting.

We train a foundation Transformer model \cite{dosovitskiy2021an} on our collected mmWave dataset for people counting, where individuals remain stationary. We compare our results against a state-of-the-art moving people counting method, as shown in Table \ref{tab:sota}. Vaidya et al. \cite{marco2024mmwave} achieved 69\% recall in an unseen environment where individuals were moving, whereas our trained model attains only 25\% recall when individuals are stationary. 

Notably, while \cite{marco2024mmwave} used a single-chip radar for moving people counting, we employ a higher-resolution four-chip cascaded mmWave radar. However, despite using enhanced hardware, we did not achieve comparable accuracy. This experiment highlights the limitations of plain spectrograms for static people counting and emphasizes the need for more advanced approaches.

\begin{table}[!t]
\centering
\begin{tabular}{|c|c|c|c|c|} \hline

\textbf{Method}
&
\textbf{Density}
&
\textbf{Body Motion}
&
\textbf{Accuracy}
\\ \hline

Vaidya et al. \cite{marco2024mmwave}
&
Dense
&
Moving
&
69\%
\\ \hline

ViT-only
&
Dense
&
Static
&
25\%
\\ \hline 

\end{tabular}
\caption{State-of-the-art people counting results}
\label{tab:sota}
\vspace{-6mm}
\end{table}

\begin{figure*}[!t]
\centering
\subfloat[Ground truth]
{\includegraphics[width=0.3\textwidth]{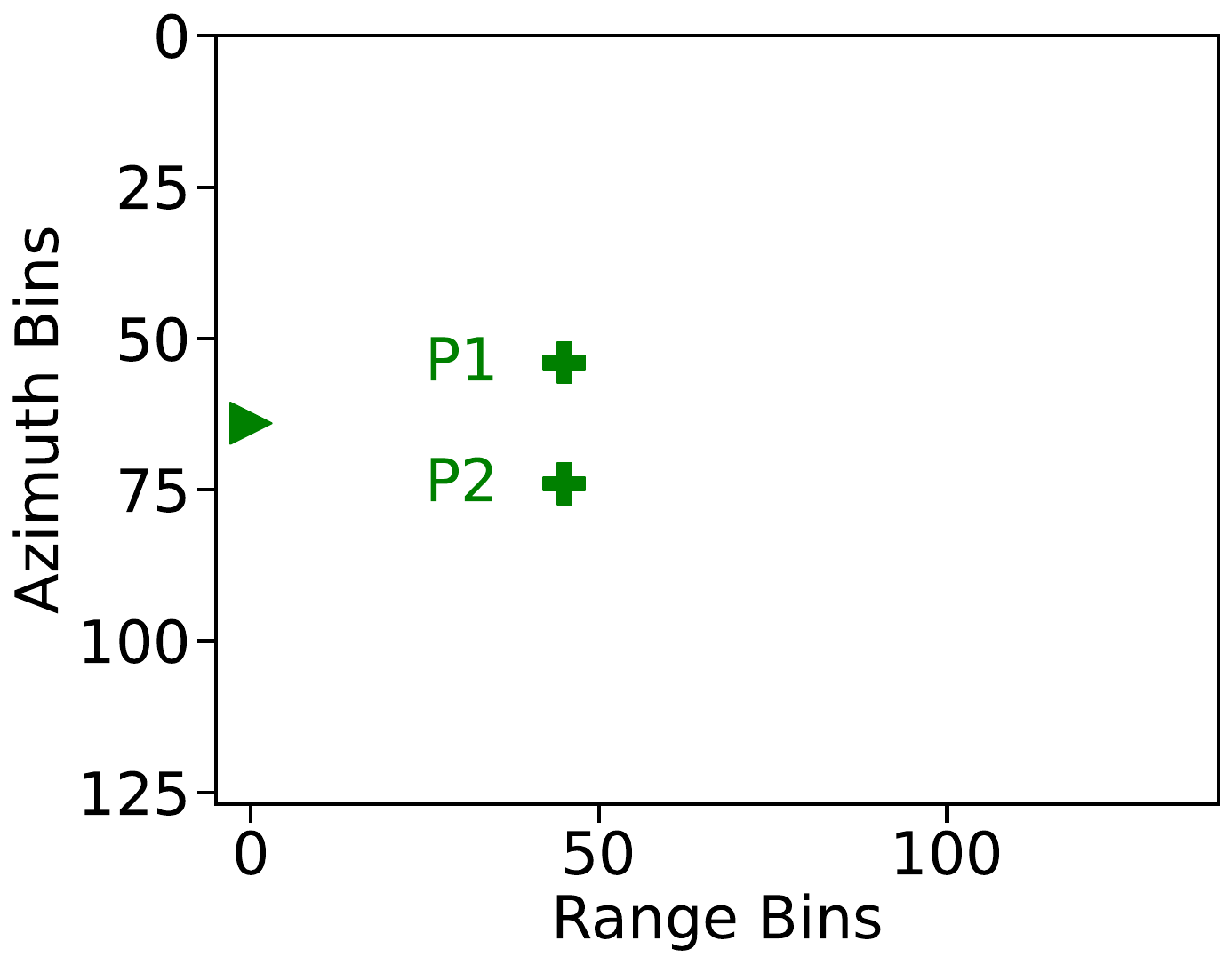}
\label{fig:gt}}
\hfil
\subfloat[Range-Azimuth spectrogram]
{\includegraphics[width=0.3\textwidth]{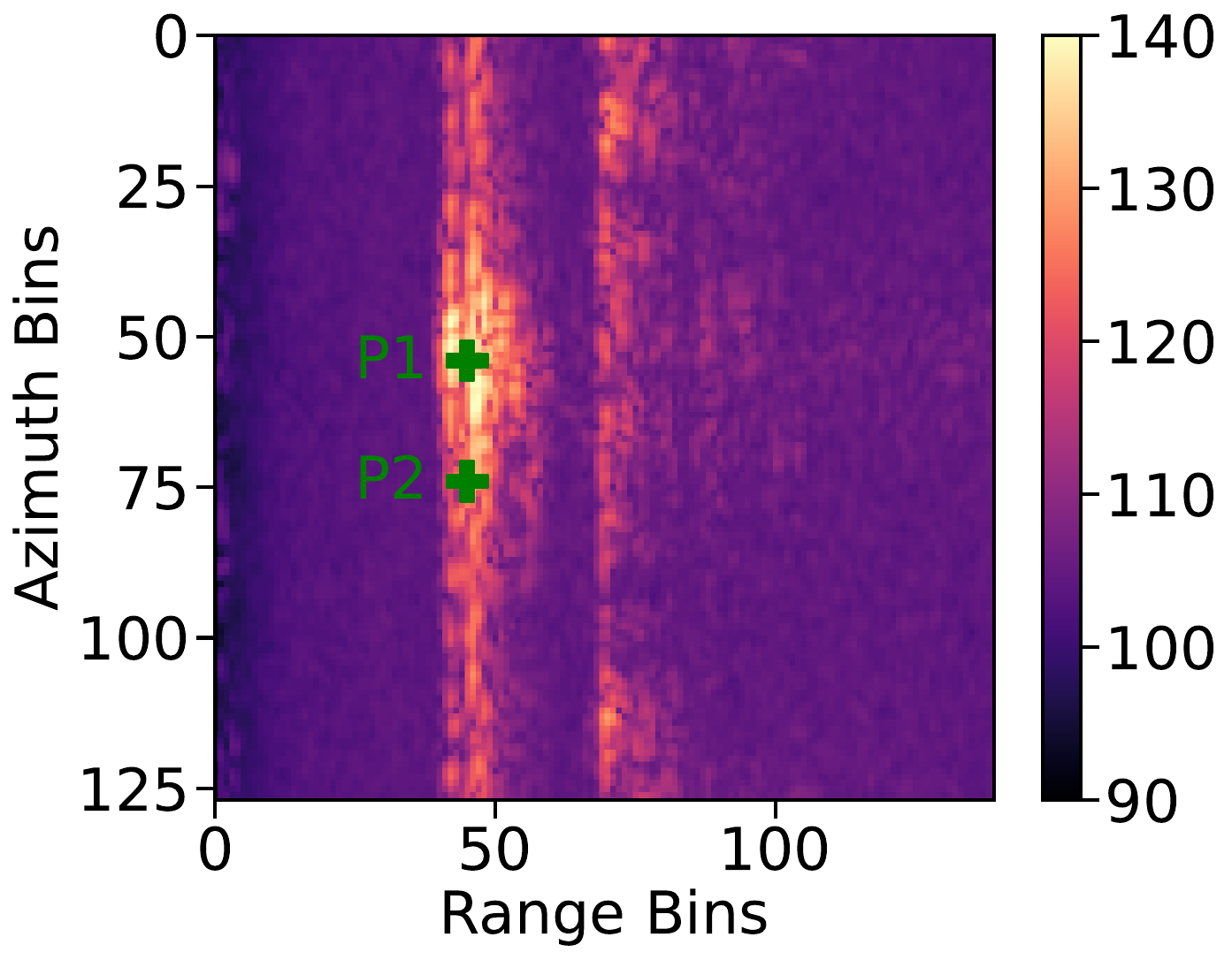}
\label{fig:fft}}
\hfil
\subfloat[Zero-elevation point cloud]
{\includegraphics[width=0.3\textwidth]{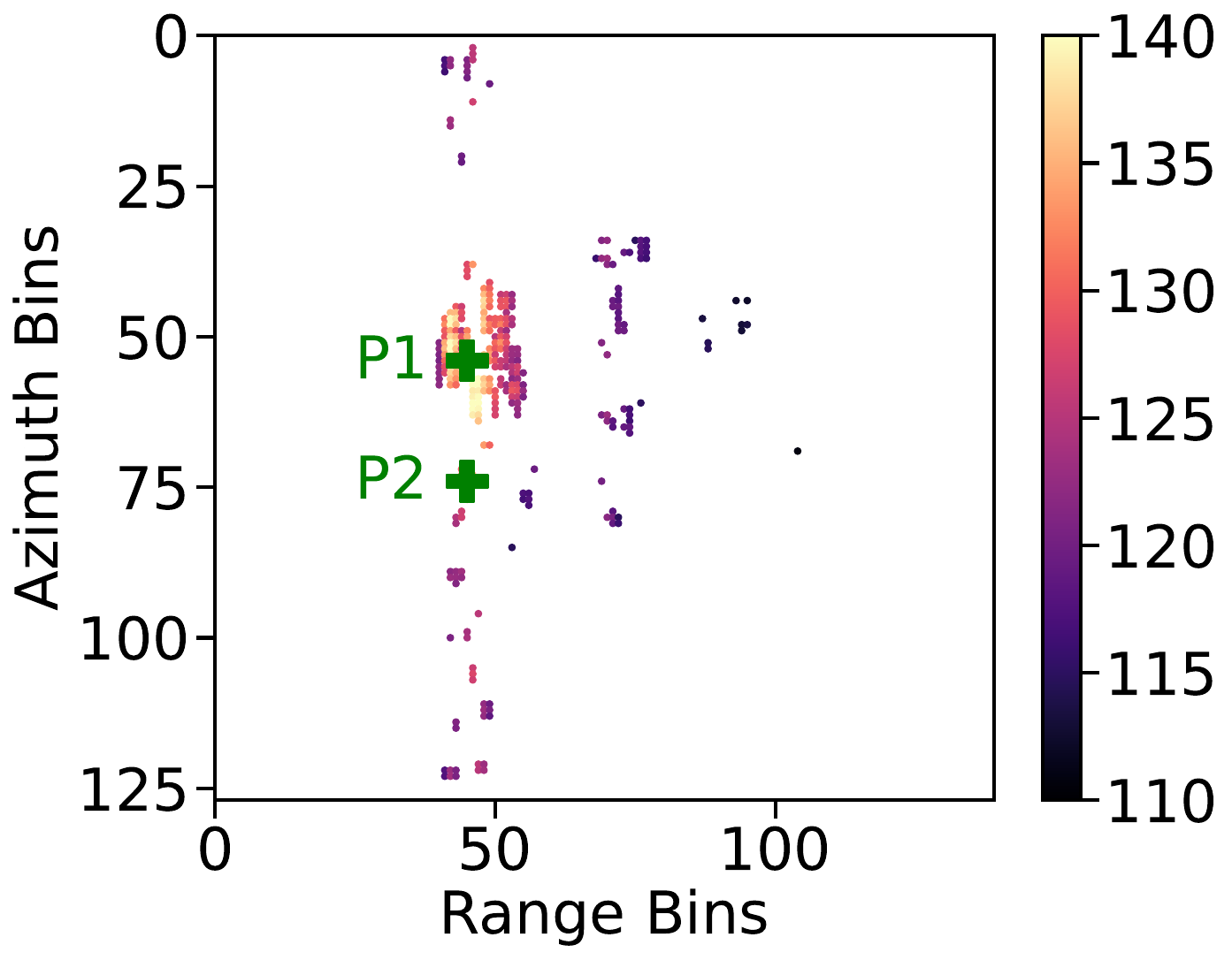}
\label{fig:pts}}
\vspace{-2mm}
\hfil
\subfloat[Micro-displacements (Range-Azimuth plane)]
{\includegraphics[width=0.3\textwidth]{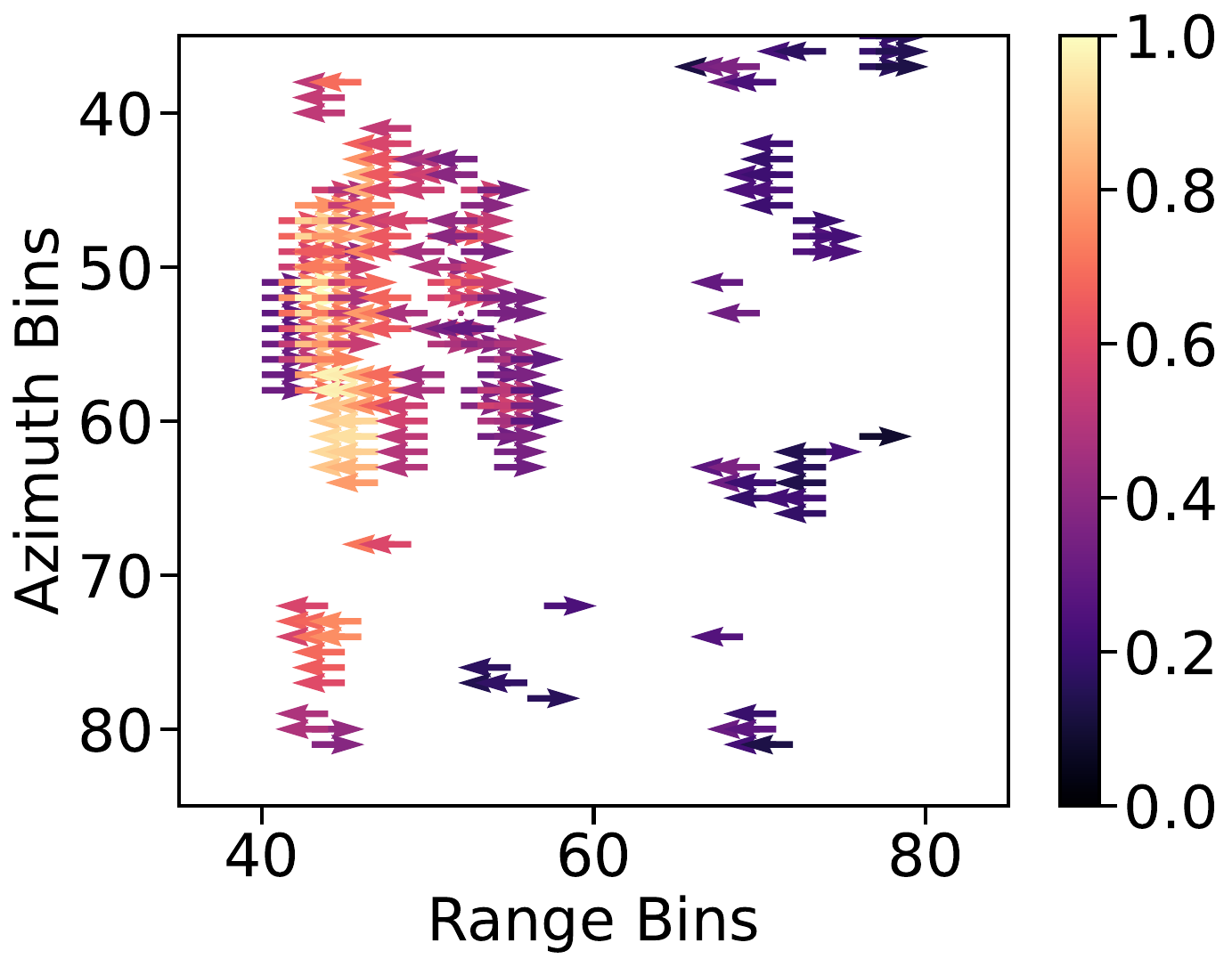}
\label{fig:micro}}
\hfil
\subfloat[Micro-displacements (over frames)]
{\includegraphics[width=0.3\textwidth]{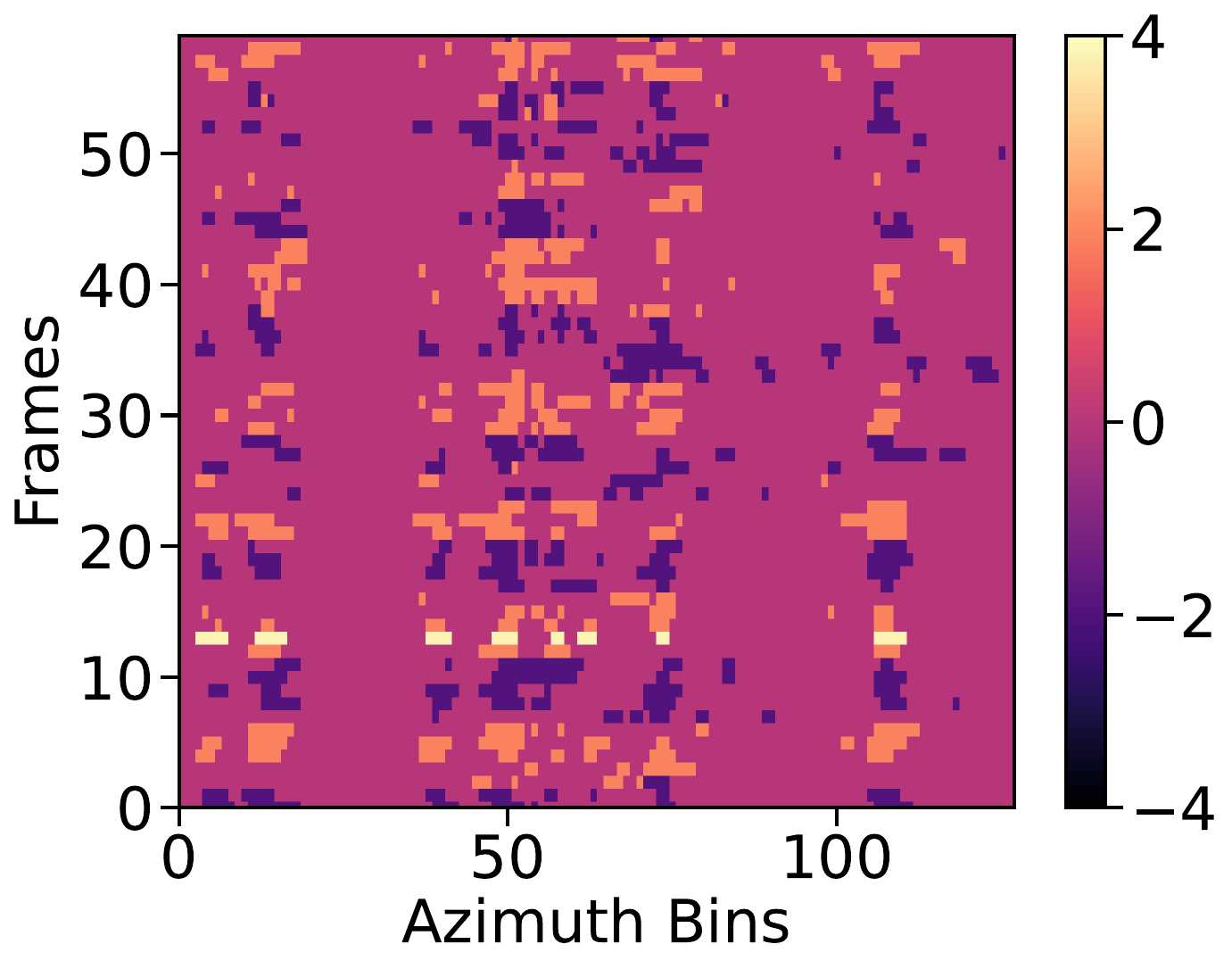}
\label{fig:frame}}
\hfil
\subfloat[Micro-motion signals (over time)]
{\includegraphics[width=0.3\textwidth]{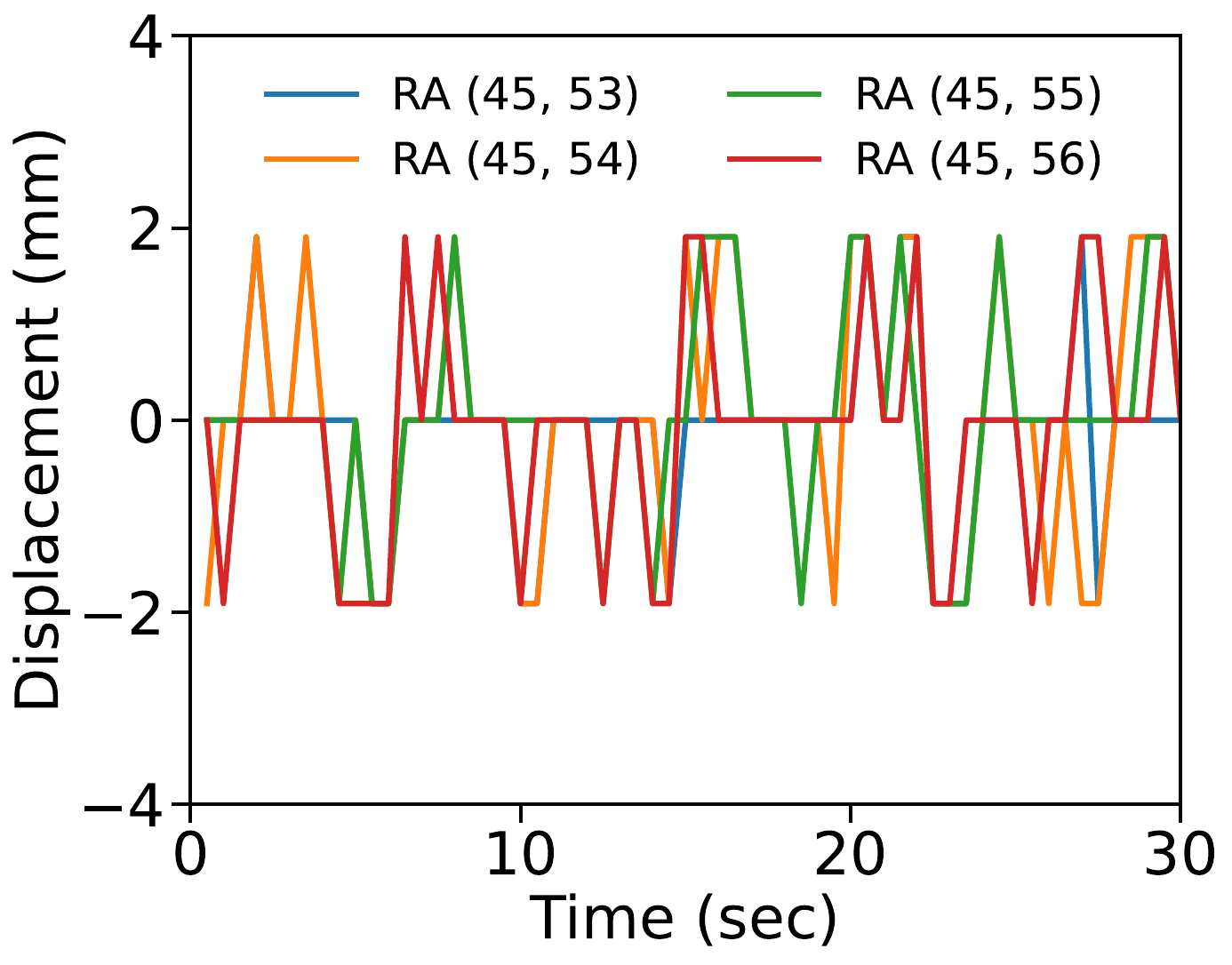}
\label{fig:signal}}
\vspace{-2mm}
\caption{Micro-motion signal extraction in a radar scene with two individuals standing side by side, $\Delta y = 6$ inches}
\vspace{-2mm}
\end{figure*}

\section{Overview of \sys}
\sys is a novel privacy-preserving people-counting system that estimates stationary and densely packed individuals using mmWave radar. It captures micro-motion signals, extracts spatial breathing patterns, and identifies unique breathing patterns to determine the number of people.  
 
\subsection{Assumptions}

\rnote{\sys operates under the following assumptions to accurately count stationary individuals in dense indoor environments within its field of view (FoV):} 

\rnote{First, the body pose of individuals influences \sys's ability to detect breathing signals, which in turn affects overall counting accuracy. When individuals face the radar (i.e., their chest is in the radar’s line of sight), detection is more reliable. However, those standing sideways or facing away (180\textdegree~back) weaken the breathing signal, potentially reducing \sys's accuracy. While deploying multiple radars could mitigate this limitation, this paper focuses on a single-radar setup.}
    
\rnote{Second, \sys relies on micro-motions of the body when individuals are mostly stationary (i.e., not walking or moving significantly) to isolate breathing signals. To ensure accurate counting, the system requires signals that are free from motion-related artifacts, such as Doppler shifts introduced by movement. As long as the captured signals contain segments where individuals remain stationary, \sys can effectively perform counting. In mixed static-dynamic scenarios, dynamic segments must be filtered out before processing.}  

\vspace{-2mm}
\subsection{System Design}
\sys consists of three core subsystems, each addressing specific challenges in detecting and counting individuals based on their micro-motions. As illustrated in Figure \ref{fig:blocks}, the system integrates new components and existing methods applied in novel contexts, with the newly proposed modules highlighted in blue.

The first subsystem extracts micro-motion signals from the raw IQ radar data, focusing on capturing subtle physiological movements such as breathing and slight body sway. This step combines established signal processing techniques with novel adaptations to enhance the detection of static individuals' micro-movements. Section~\ref{sec:step1} describes this in detail. 

The second subsystem isolates the breathing components from these micro-motion signals. This stage separates the periodic breathing signals, which are key for identifying individuals, from other random body motions. While some signal decomposition methods used here are conventional, they are applied in a novel context to handle multiple people in a dense environment. Section~\ref{sec:step2} describes this in detail.

The third subsystem groups the breathing components based on spatial information, determining the number of individuals present by identifying distinct clusters of breathing signals. This spatial grouping module is a unique contribution of our system, tailored specifically to the challenge of mapping multiple breathing signals to individuals in static, crowded settings. Section~\ref{sec:step3} describes this in detail.

\section{Micro-Motion Signal Extraction}
\label{sec:step1}

This module processes the raw IQ signals captured by the mmWave radar, which transmits equidistant chirp signals via multiple transmitter antennas. These chirps reflect off human body parts and are received by the radar through ADC sampling. The raw IQ signals are then input into the FFT module for further analysis.

For example, consider a scenario where two individuals, P1 (45, 54) and P2 (45, 74), stand side-by-side with minimal spatial separation of 6 inches, as shown in Figure~\ref{fig:gt}. This demonstrates how the radar captures the scene and how closely spaced individuals are reflected in the signal.



\vspace{-1mm}
\subsection{Range-Doppler-Azimuth FFT}

We first localize individuals in the radar data to extract micro-motion. We apply FFT three times to the raw IQ signals along the sample, chirp, and antenna axes. This produces the Range-Azimuth frequency spectrogram, as shown in Figure~\ref{fig:fft}, where brighter pixels (stronger reflections) indicate higher motion from human body parts.

\vspace{-1mm}
\subsection{Zero-Elevation Point Cloud}
The Range-Azimuth spectrogram contains dense spatial information that must be filtered to isolate object points. We apply the Cell-Averaging Constant False Alarm Rate (CFAR) algorithm \cite{hm1968adaptive} to remove noise, leaving a point cloud, as shown in Figure~\ref{fig:pts}. Some additional (ghost) points appear far from the standing individuals due to multi-path interference, common in indoor environments \cite{yue2018extracting}. Notably, we exclude elevation data to simplify analysis and avoid the complexity of three-dimensional processing.  


\begin{figure*}[!t]
\centering
\subfloat[Micro-source signals]
{\includegraphics[width=0.3\textwidth]{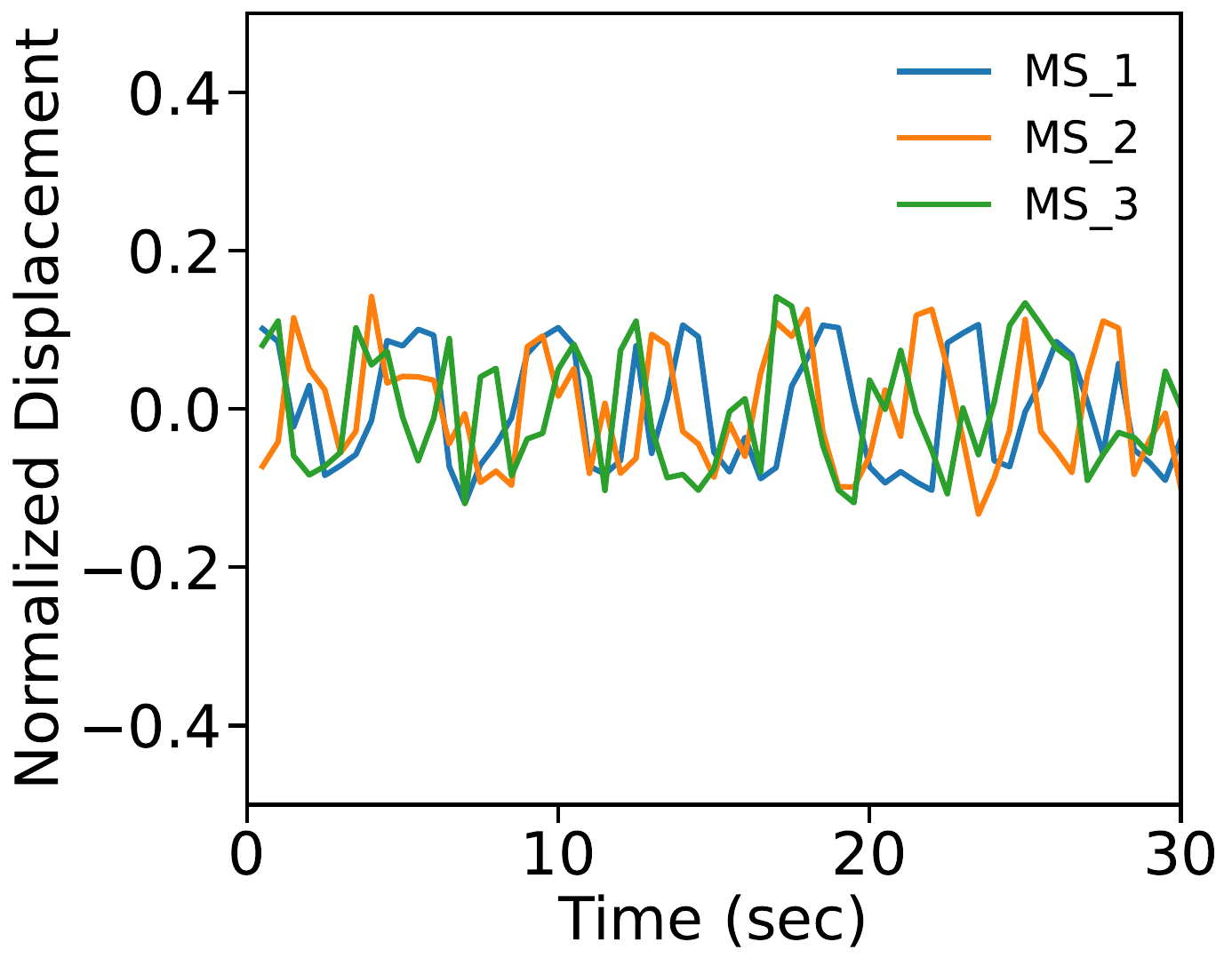}
\label{fig:sources}}
\hfil
\subfloat[Frequency spectrogram of source signals]
{\includegraphics[width=0.3\textwidth]{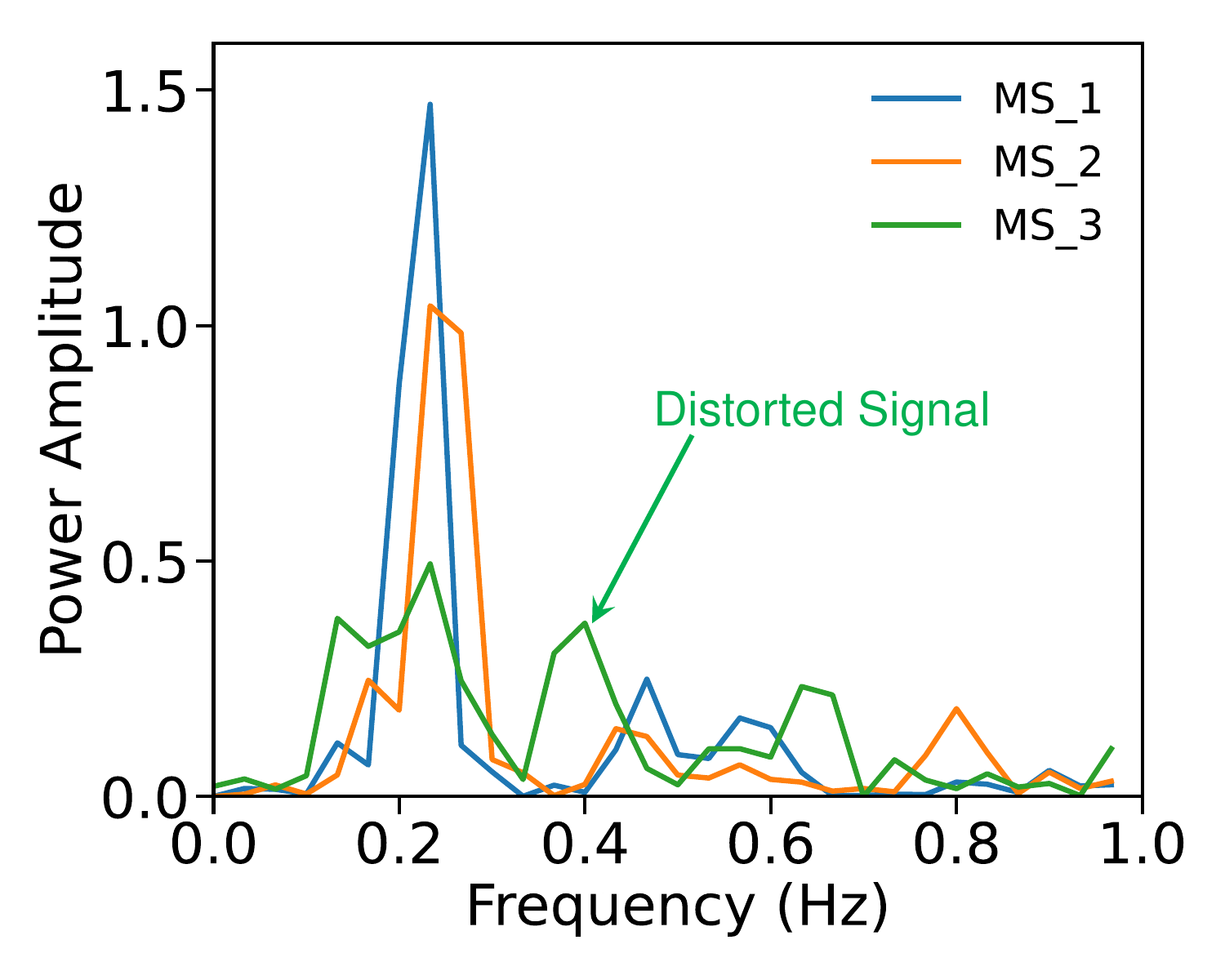}
\label{fig:freq}}
\hfil
\subfloat[Breathing signals]
{\includegraphics[width=0.3\textwidth]{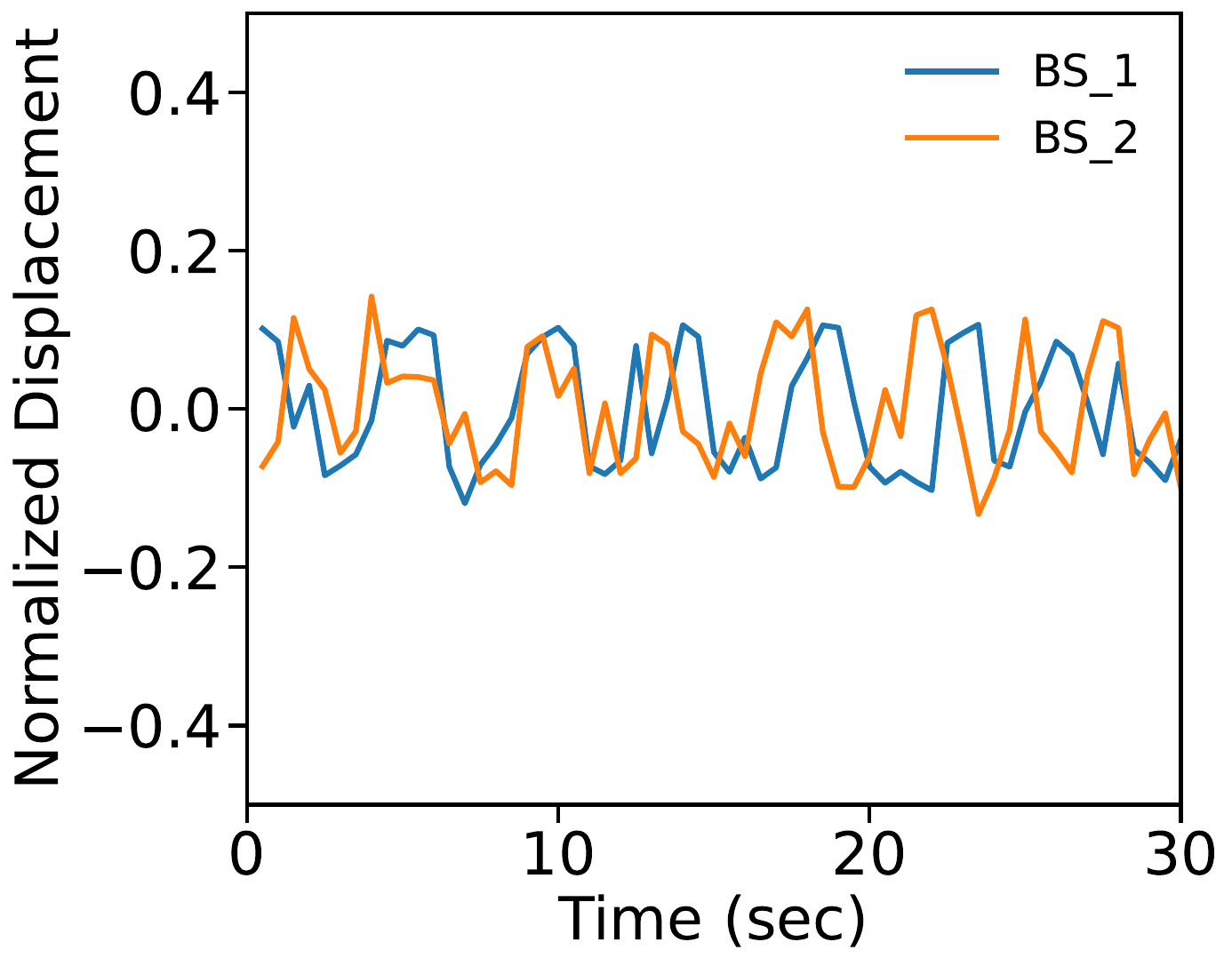}
\label{fig:breathing}}
\vspace{-2mm}
\hfil
\subfloat[Spatial mapping of BS\_1]
{\includegraphics[width=0.3\textwidth]{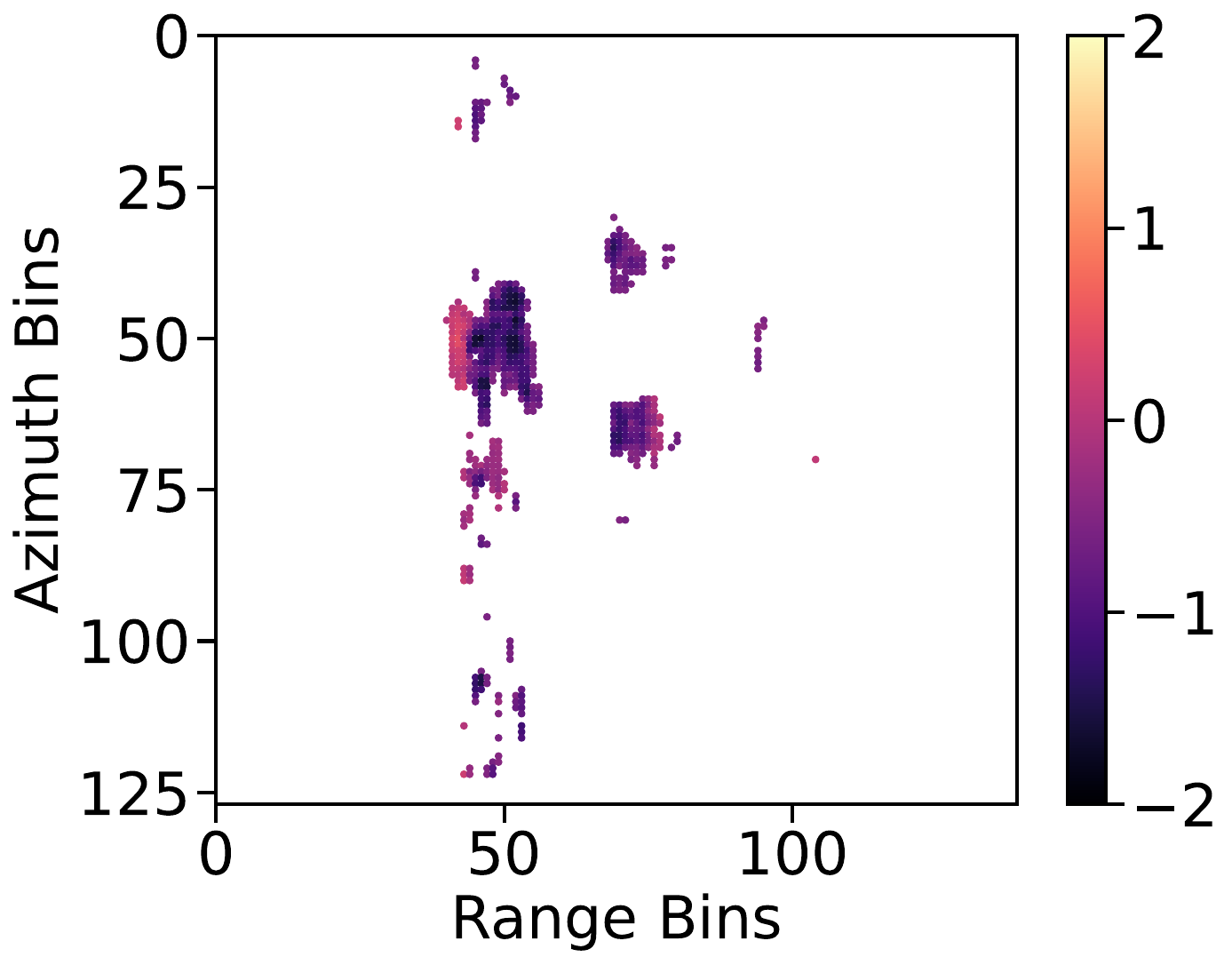}
\label{fig:weight1}}
\hfil
\subfloat[Spatial mapping of BS\_2]
{\includegraphics[width=0.3\textwidth]{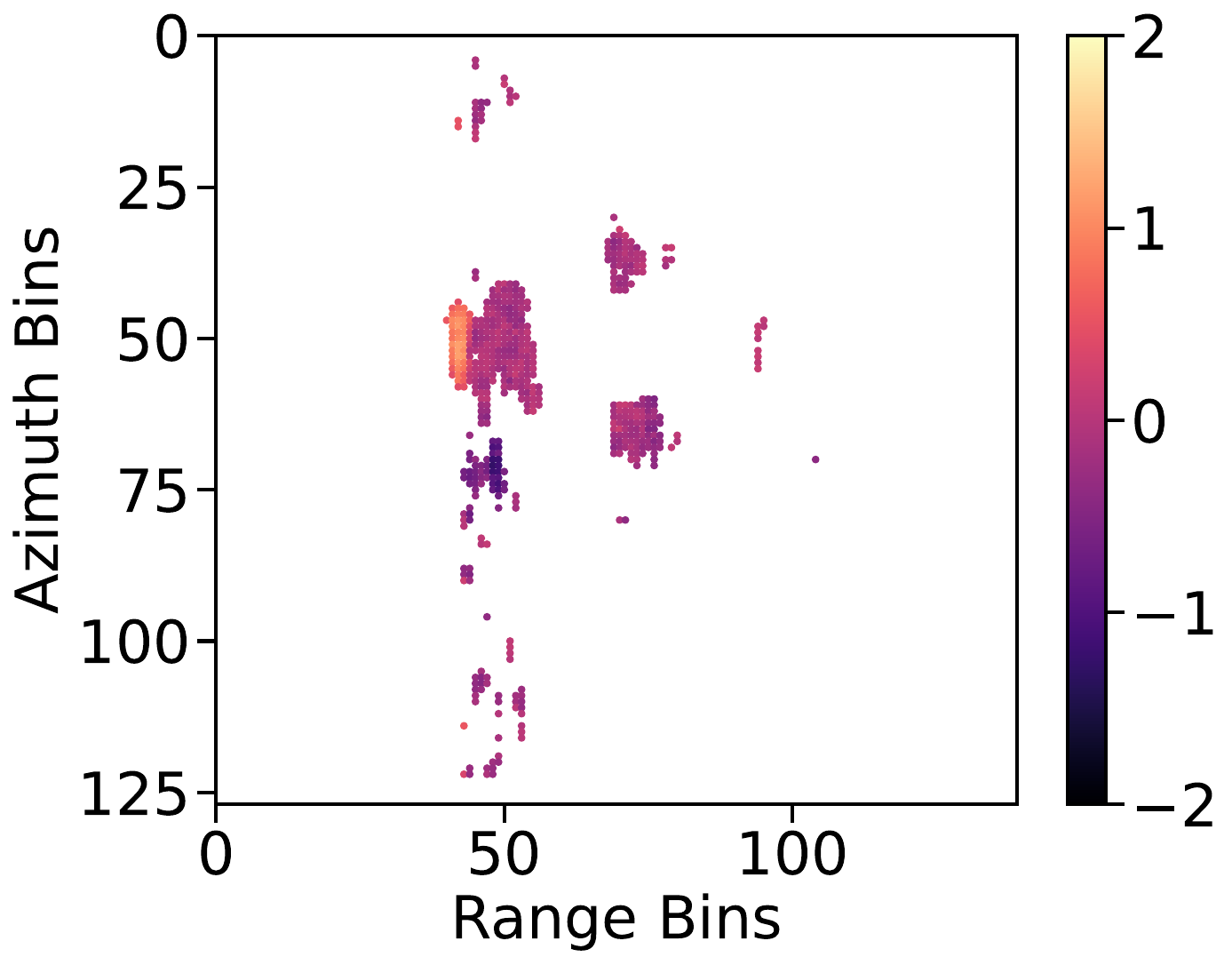}
\label{fig:weight2}}
\hfil
\subfloat[Location of the breathing sources]
{\includegraphics[width=0.3\textwidth]{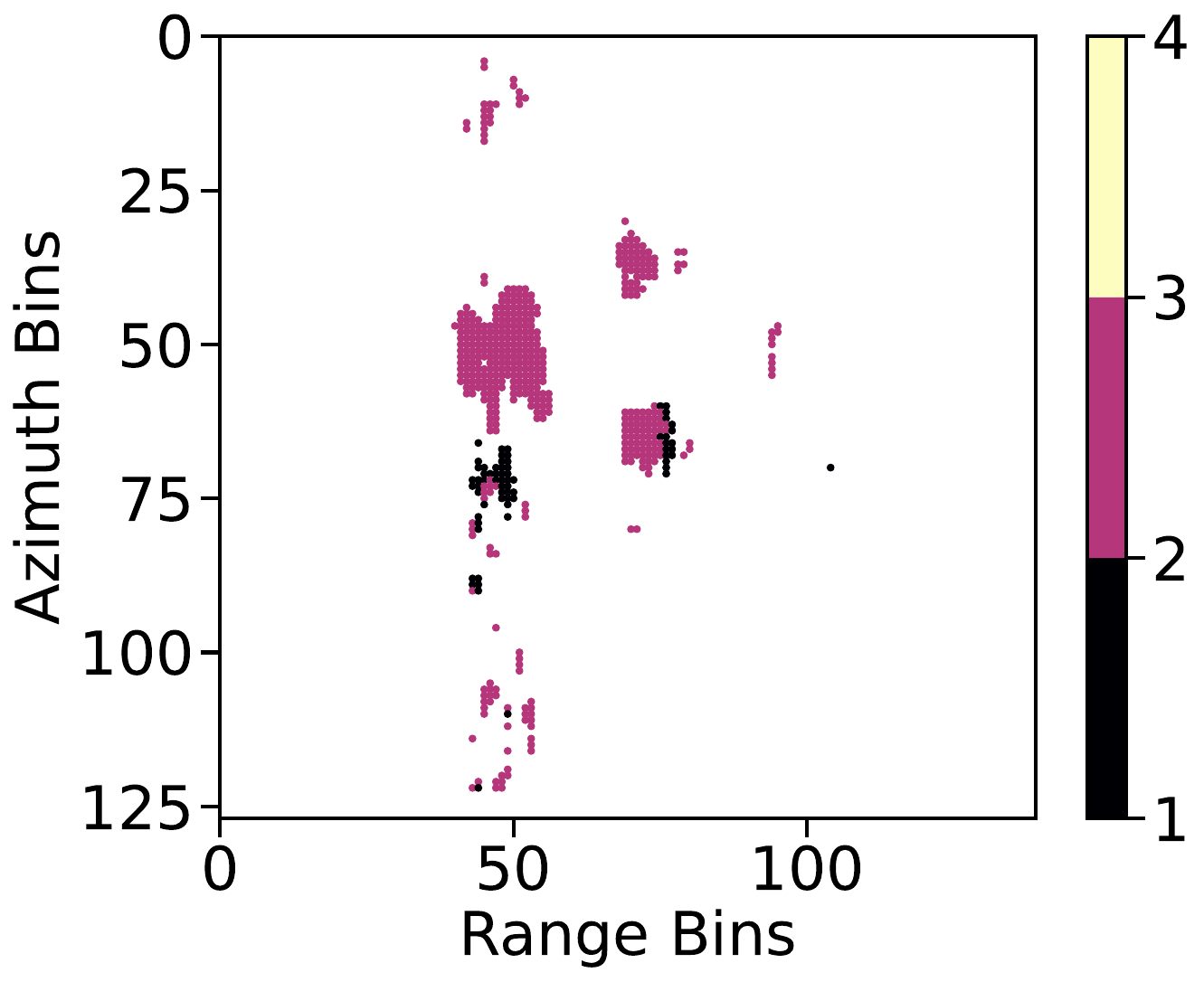}
\label{fig:weight}}
\vspace{-2mm}
\caption{Breathing component separation in a radar scene with two individuals standing side by side, $\Delta y = 6$ inches}
\vspace{-2mm}
\end{figure*}

\subsection{Micro-Displacement Estimation}
The zero-elevation point cloud cannot distinguish between two static individuals. To overcome this, we extract the natural micro-motion signals from human body parts by estimating the micro-displacement values of object points using their Doppler information.

A mmWave radar sends a set of chirps at regular intervals, forming a frame. If the peak Doppler velocity of the $i$-th radar point $R_i$ at the $j$-th frame is $v_{ij}$, the displacement is given by $d_{ij} = v_{ij} \times T_F$, where $T_F$ is the frame duration. We use this formula to estimate displacement values for each frame, as shown in the zoomed quiver plot in Figure~\ref{fig:micro}. Here, radar points indicate two types of displacements: either approaching or moving away from the radar, as it detects only positive or negative velocity within a single frame.

\rnote{By estimating the micro-displacements across several frames, we can observe the body movement of an individual.} Figure \ref{fig:frame} displays micro-displacements from a recording, with yellow pixels indicating positive displacements (moving away) and dark pixels showing negative displacements (approaching). Some points also move along the Azimuth direction, indicating body swaying.

These displacements over time constitute human micro-motion signals. Figure~\ref{fig:signal} illustrates signals from four radar points, sampled every 0.5 seconds. Since displacements are discrete, they appear smaller than actual chest movements. For instance, inhaling over 3 seconds results in cumulative micro-displacements exceeding 1 cm.  

In this way, we extract human micro-motion from raw IQ radar signals in terms of micro-displacements through our robust signal processing techniques. Next, we discuss how we extract the breathing sources from these micro-motion signals.

\section{Breathing Component Separation}
\label{sec:step2}

The recorded micro-motion signals arise from human breathing and other micro-activities. Due to multi-path interference, one person’s breathing signal may mix with another’s, requiring the decomposition of the mixed signals into their sources. We explain this process in this section.

\subsection{Invalid Micro-Motion Removal}

During the recording session, a person may sway their body, causing changes in the number or location of radar points. As a result, not all radar points provide reliable micro-motion signals. A valid signal must (a) have at least $m$\% non-zero displacement values and (b) include both positive and negative displacement values. Based on these criteria, we filter out all invalid signals and their corresponding radar points.

\subsection{Micro-Source Separation}
The micro-motion signal is a mixture of multiple source signals from different individuals, forming a linear combination of independent, non-Gaussian signals \cite{yue2018extracting}. To separate these signals, we apply the Independent Component Analysis (ICA) algorithm \cite{hyvarinen2000independent}. However, ICA requires the number of source signals, which is unknown. To address this, we iteratively run ICA, increasing the number of components from $i=1, 2, ..., n$. In each iteration, we extract $i$ components and their spatial mappings, generating a total of $n(n+1)/2$ micro-source signals. Figure \ref{fig:sources} illustrates three decomposed source signals, where MS\_i denotes the $i$-th micro-source signal. These signals exhibit periodic patterns characteristic of breathing.

\begin{figure*}[!t]
\centering
\subfloat[Spatial breathing profile]
{\includegraphics[width=0.3\textwidth]{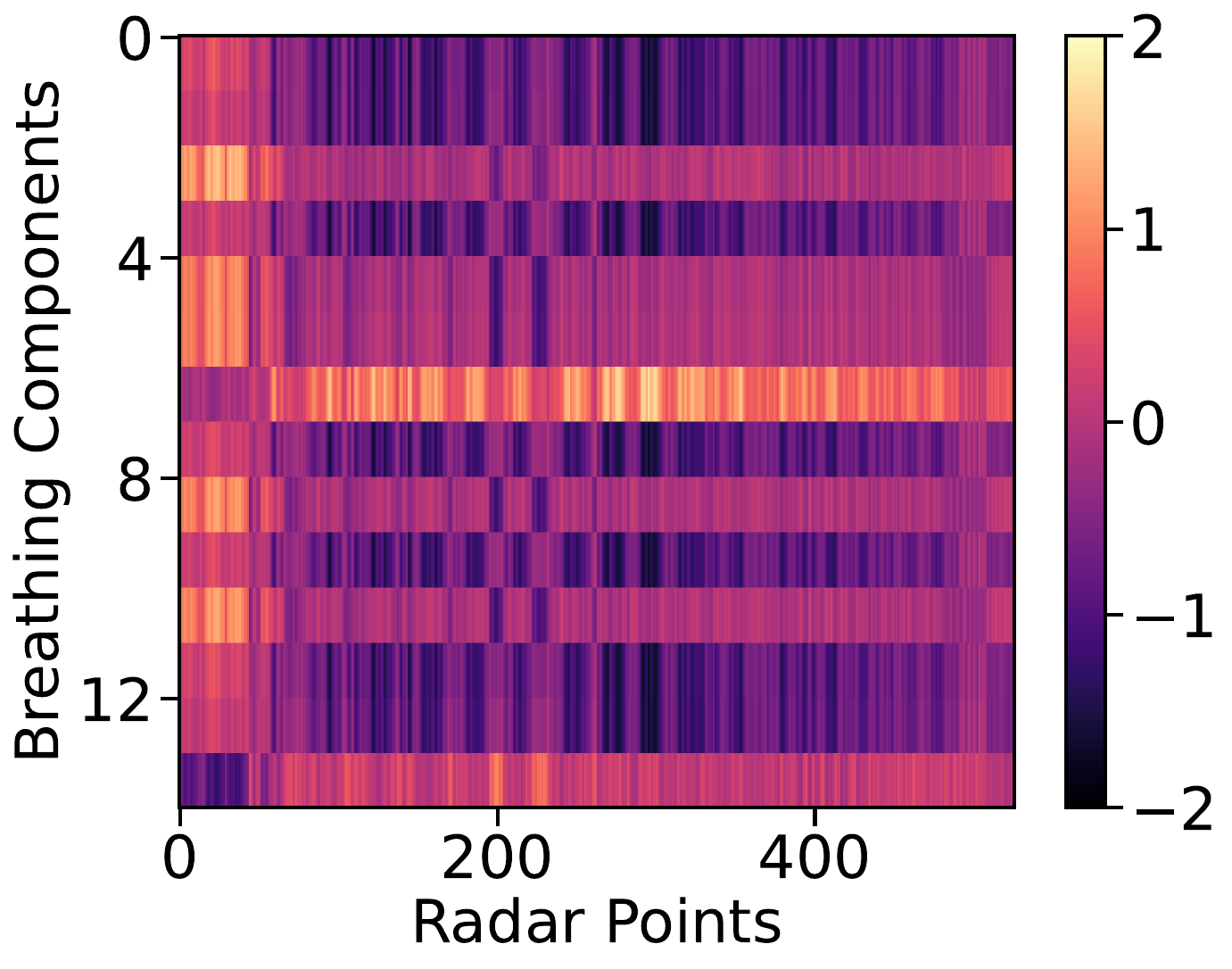}
\label{fig:profile}}
\hfil
\subfloat[Dendrogram of the breathing profile]
{\includegraphics[width=0.3\textwidth]{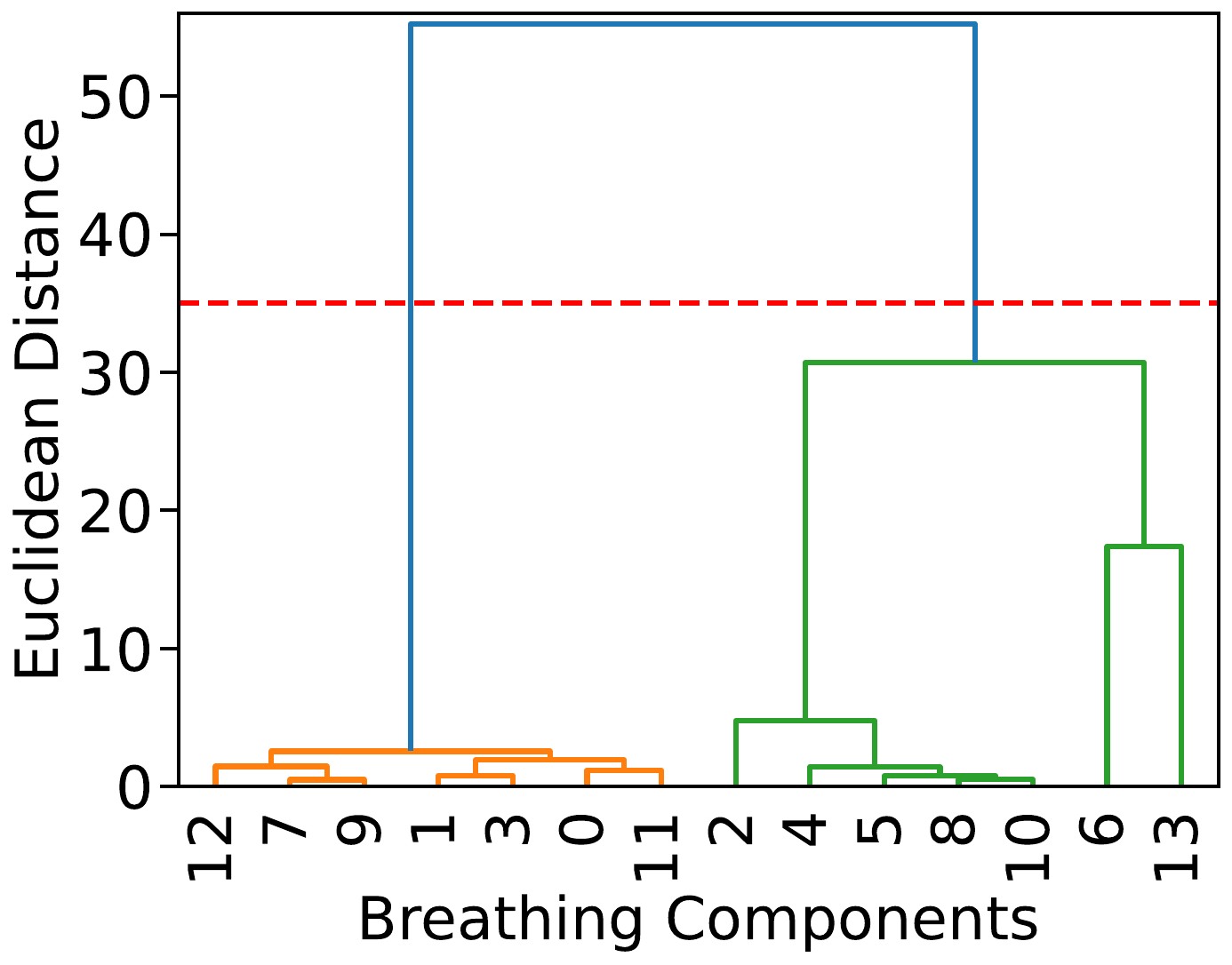}
\label{fig:cluster}}
\hfil
\subfloat[Breathing profile learning]
{\includegraphics[width=0.3\textwidth]{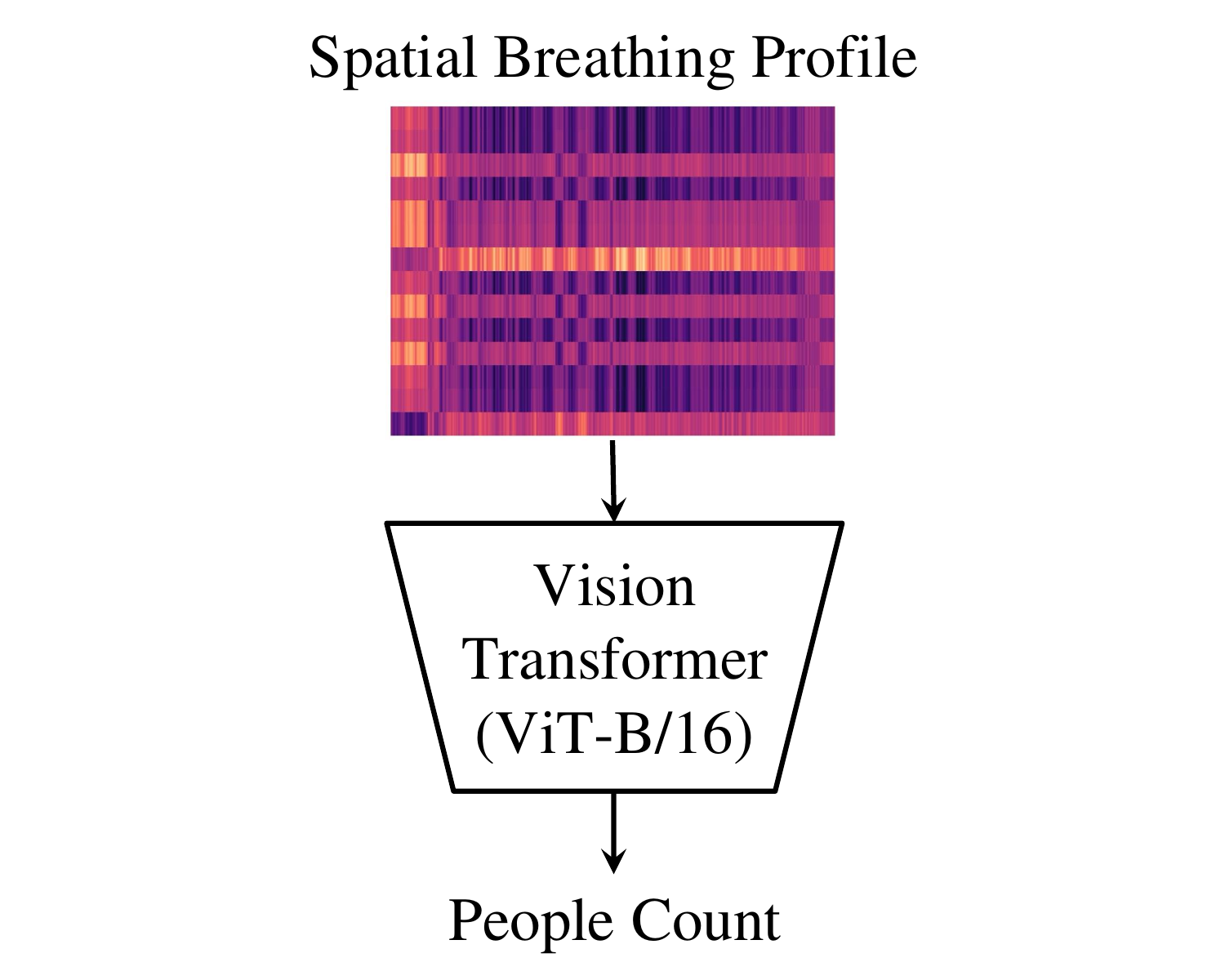}
\label{fig:model}}
\vspace{-2mm}
\caption{Breathing source grouping in a radar scene with two individuals standing side by side, $\Delta y = 6$ inches}
\vspace{-2mm}
\end{figure*}

\subsection{Breathing Source Extraction}

After applying ICA, we obtain various source signals, some of which are non-breathing signals. To filter out the non-breathing signals, we analyze the frequency spectrogram of each source. Figure~\ref{fig:freq} shows the spectrograms of three source signals, where the first two appear to be valid breathing signals, unlike the third. A source signal is considered a breathing signal if it satisfies two conditions:
\begin{align}
    b_l \leq \Bar{f} \leq b_h &\implies b_l \leq \frac{\sum_i P_i f_i}{\sum_i P_i} \leq b_h \label{eq:br} \\
    q_s \geq b_s &\implies \frac{\max_i (P_i)}{\sum_i P_i} \geq b_s \label{eq:bs}
\end{align}
Here, $f_i$ represents the frequency at bin $b_i$, and $P_i$ is the corresponding power measurement. The mean frequency of the breathing signal must fall within the breathing frequency band $[b_l, b_h]$, and the signal quality $q_s$ must exceed a breathing score $b_s$.

This process decomposes the micro-motion signal into its constituent breathing signals, as shown in Figure~\ref{fig:breathing}, where BS\_i represents the $i$-th breathing signal. Figures~\ref{fig:weight1} and~\ref{fig:weight2} display the spatial mappings of BS\_1 and BS\_2, respectively, indicating radar points' positive and negative contributions to each signal. The combined spatial mapping in Figure \ref{fig:weight} shows that BS\_1 originates from P1 and BS\_2 from P2. 

\rnote{The accuracy of the breathing source extraction algorithm depends on the number of frames used. A higher number of radar frames means a higher number of observations, which in turn improves the accuracy of source separation.}

\section{Breathing Source Grouping}
\label{sec:step3}


At this stage, we have a set of breathing signals and need to determine how many individuals generated them. Due to multiple ICA iterations, a single person's breathing signal may appear multiple times, resulting in a many-to-one correspondence. To resolve this, we must group the breathing signals to identify the number of individuals. This section details this grouping process.

\subsection{Spatial Breathing Profile}
We utilize the spatial mapping of breathing signals for people counting. If two breathing signals share similar spatial mappings, they likely originate from the same individual. To represent this, we stack the spatial mappings of all breathing signals, forming a \emph{spatial breathing profile}. Figure \ref{fig:profile} illustrates an example of a spatial breathing profile for a radar scene, where ICA was applied ten times, yielding 14 breathing components. Most components cluster into two groups, indicating the presence of two static individuals. \rnote{Notably, we do not assume that different individuals have distinct breathing rates. Instead, our approach focuses on the spatial locations of breathing signals rather than their rates.} Next, we discuss automated methods for grouping these components.

\subsection{Learning Breathing Groups}
The Agglomerative Clustering algorithm is one approach for automating breathing source grouping, as it hierarchically clusters data based on similarity. We apply this method to our spatial breathing profile, generating the dendrogram shown in Figure~\ref{fig:cluster}. Visually, cutting the dendrogram at 35 Euclidean distance units would yield two clusters. However, in practice, this manual approach is not feasible, and no automated method reliably determines the optimal cutoff distance. Therefore, we adopt a supervised approach for grouping breathing components. 

We utilize a foundation model (Vision Transformer) \cite{dosovitskiy2021an} to learn breathing groups from spatial breathing profiles. The training pipeline is illustrated in Figure~\ref{fig:model}. First, we divide the breathing profiles into patches and input them into a self-attention-based Transformer encoder to extract latent features. Finally, these features are mapped into people count. Note that, we do not use any RGB camera images. Instead, we generate the spatial breathing profiles from mmWave signals and input them into the model to get the people count.


\rnote{We select target classes based on variations in density, including two low-density cases (2 and 3 people per 3 m$^2$) and two high-density cases (5 and 7 people per 3 m$^2$). Intermediate group sizes, such as 4 and 6, also fall under high-density conditions and are therefore excluded. This exclusion does not introduce any fundamental limitations to our system. In addition, we do not consider group sizes of 0 and 1, as detecting human presence in such cases is a well-studied problem \cite{marco2024mmwave}.}

\begin{figure*}[!t]
\centering
\subfloat[Radar setup]
{\includegraphics[width=0.3\textwidth]{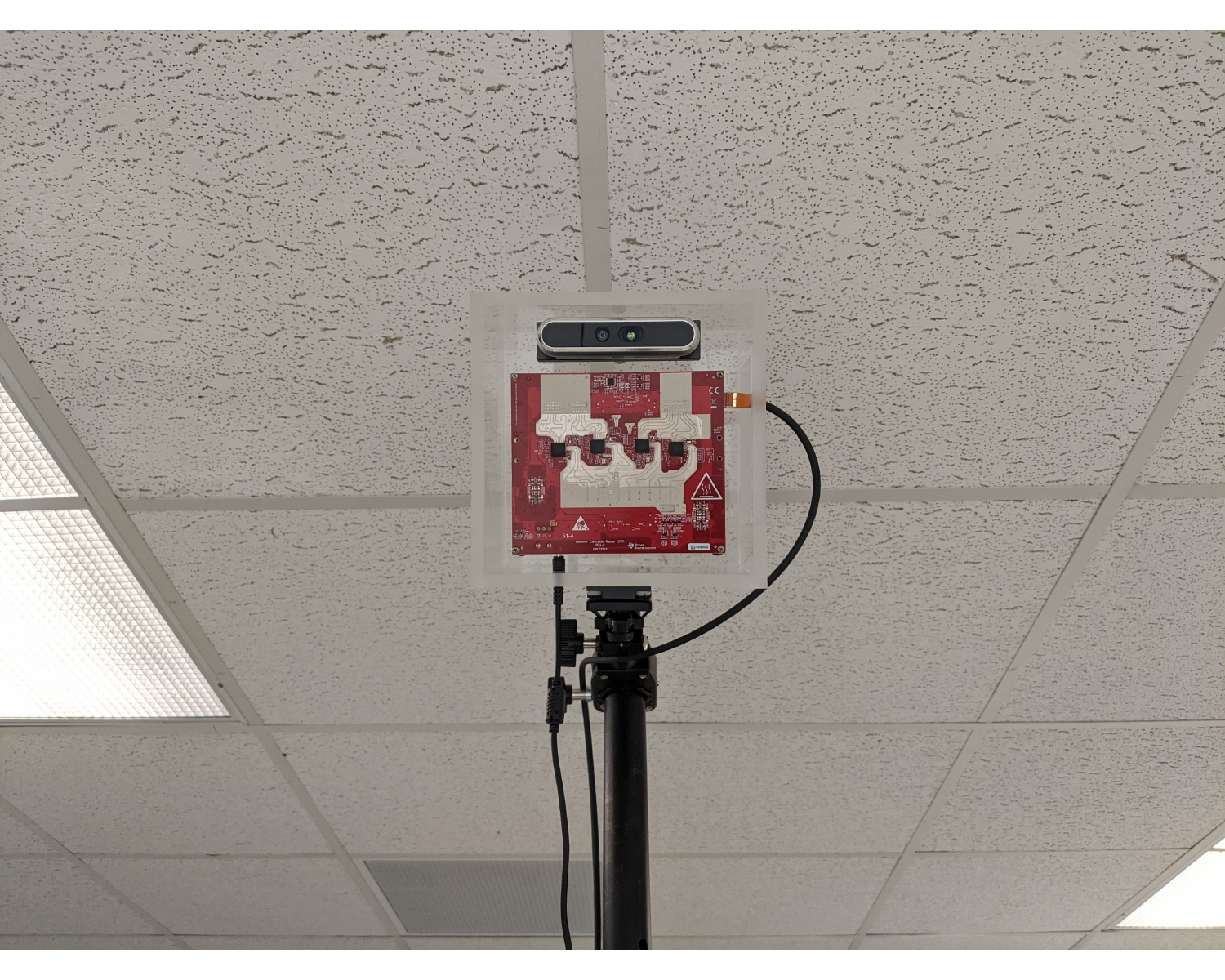}
\label{fig:radar}}
\hfil
\subfloat[Data collection in lab (D1, D2)]
{\includegraphics[width=0.3\textwidth]{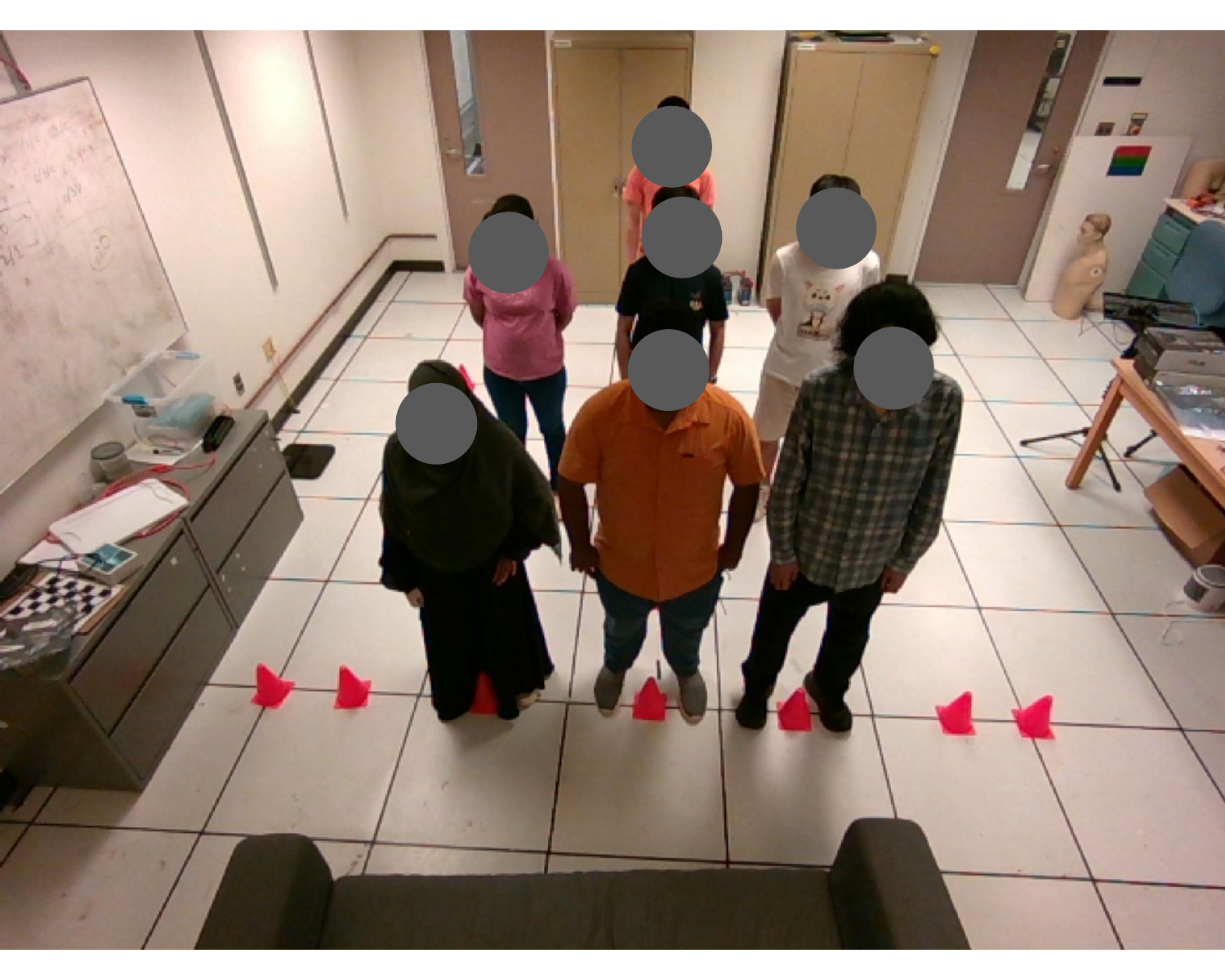}
\label{fig:lab_data}}
\hfil
\subfloat[Data collection in corridor (D3)]
{\includegraphics[width=0.3\textwidth]{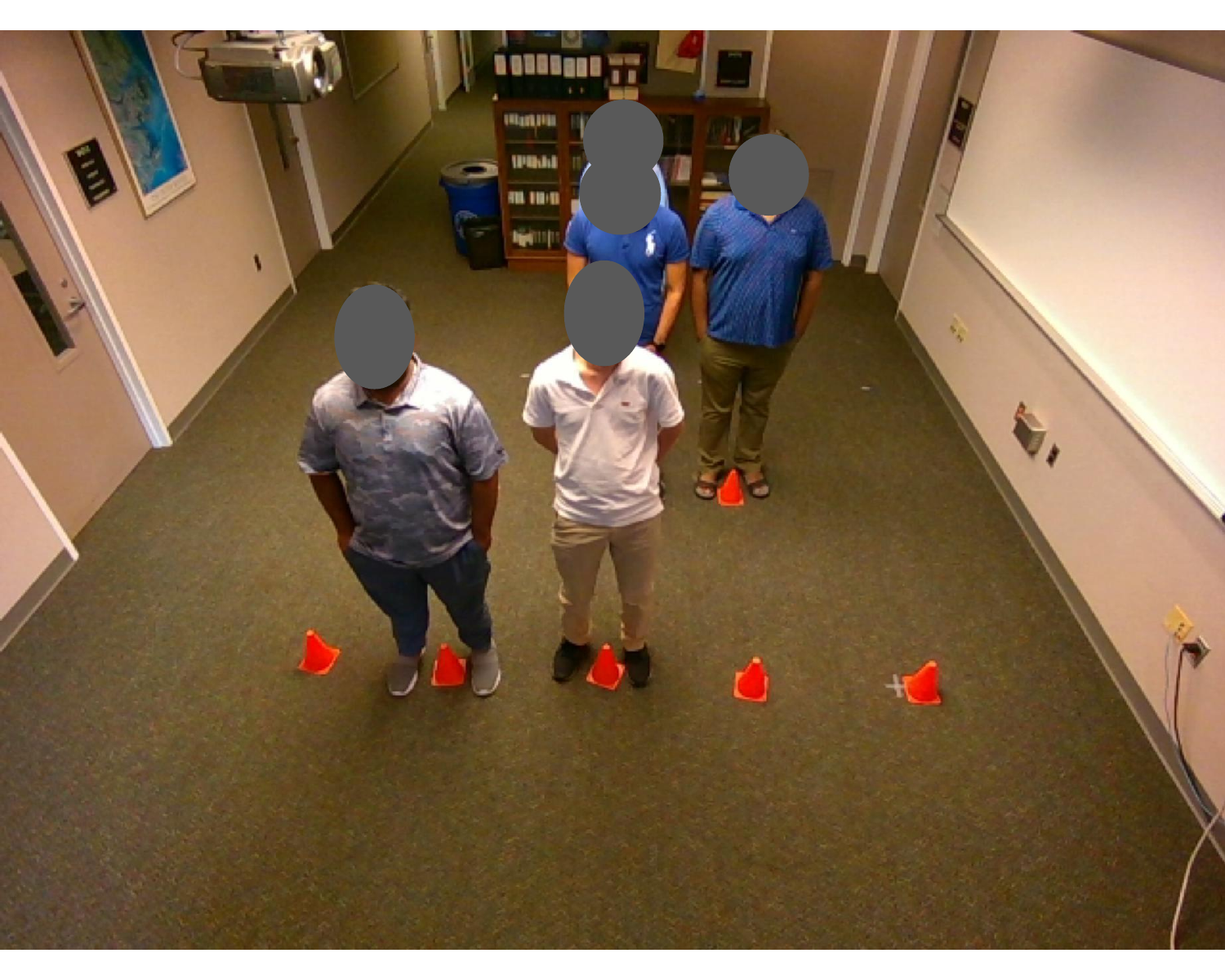}
\label{fig:lib_data}}
\vspace{-3mm}
\caption{Experimental setup}
\vspace{-3mm}
\end{figure*}

\subsection{Breathing Group Prediction}

The ViT model learns the inter-component similarity between breathing components from the spatial breathing profile during training. We then evaluate the model in a new environment with different users, and it accurately predicts the number of breathing groups. This demonstrates that our system can count the number of static individuals in dense indoor scenarios by leveraging breathing signals.

In the preliminary study (Section \ref{sec:prelim}), we trained an end-to-end ViT model on Range-Azimuth spectrograms, but their lack of identifiable signatures led to poor accuracy. This motivated our breathing-signal-based approach, which captures unique breathing signatures. 

\begin{table*}[!t]
  \centering
  \begin{tabularx}{\textwidth}{|Y|c|Y|c|Y|c|c|c|} \hline
    \textbf{Parameter} & \textbf{Value} & \textbf{Parameter} & \textbf{Value} & \textbf{Parameter} & \textbf{Value} & \textbf{Parameter} & \textbf{Value} \\ \hline
    
    Center Frequency & 79.13 GHz & ADC Samples & 256 & Range Resolution & 42.1 mm & Max Range & 10.77 m \\ \hline
    
    Wavelength & 3.79 mm & Frame Time & 89 ms & Velocity Resolution & 21.2 mm/s & Max Velocity & 1.36 m/s \\ \hline
    
    Chirp Bandwidth & 3.56 GHz & Frame Rate & 2 Hz & Azimuth Resolution & 1.4 deg & Azimuth FoV & $\pm$ 70 deg \\ \hline
    Chirp Count & 128 & Frame Count & 60 & Elevation Resolution & 18 deg & Elevation FoV & $\pm20$ deg \\ \hline
  \end{tabularx}
  \caption{Configuration parameters of our mmWave cascaded radar}
  \label{tab:param}
  \vspace{-6mm}
\end{table*}

\vspace{-2mm}
\section{Experimental Setup}

\subsection{Radar Setup}
We use a commercial off-the-shelf mmWave cascade imaging radar from Texas Instruments \cite{ti2024radar}, featuring a cascaded array of four AWR2243 devices that function as a single high-resolution RF transceiver. The radar's configuration parameters are listed in Table \ref{tab:param}. To capture fine-grained human micro-motion, such as torso displacement from breathing, we increase the number of chirps. Due to the radar's low elevation resolution, we exclude elevation data from our system pipeline. Figure \ref{fig:radar} shows the radar setup, mounted on an eight-foot tripod with a custom acrylic casing and inclined at 40 degrees to cover the experimental area. We use a Linux-based millimeter-wave capture toolkit \cite{Lu_Millimeter-wave_Capture_Standard} to record the scene.


\subsection{Data Collection}
\rnote{Due to the lack of public radar-based people-counting datasets, we collect our own by recruiting 21 participants, ensuring diversity across gender, age, height, and weight. Our dataset includes 16 male and 5 female participants, with ages ranging from 18 to 55 years, heights from 5 to 7 feet, and weights from 120 to 180 lbs.}  

To ensure comprehensive coverage from simple to complex cases and improve generalization, we systematically select representative group patterns instead of using random arrangements, which might overlook critical configurations. We vary factors such as interpersonal distance, radar-to-human distance, and radar angle. Participants are grouped into varying-sized sets, forming spatial arrangements such as side-by-side, front-to-back, triangular, diamond, and diagonal patterns. The minimum distances are $\Delta y=6$ inches for side-by-side and $\Delta x=3$ feet for front-to-back.  

We record 125 group patterns in two distinct environments, as shown in Figures \ref{fig:lab_data} and~\ref{fig:lib_data}. Data collection follows a structured approach to ensure precise measurement of distances and angles. However, this does not limit our system’s applicability—it remains effective for unstructured groups, allowing individuals to be positioned anywhere within the radar’s field of view (FoV), which is 3 m$^2$ in our experiments. 

The dataset is divided into three subsets: D1 (54 groups), D2 (51 groups), and D3 (20 groups). We train our model on D1 and test its robustness on D2, both collected in the first environment, while D3 evaluates performance in a second environment.

\vspace{-3mm}
\subsection{Breathing Profile Generation}
We apply signal-processing algorithms, including FFT \cite{cooley1965algorithm}, CA-CFAR \cite{pan2019openradar}, and FastICA \cite{hyvarinen2000independent}, to generate the spatial breathing profile. Parameters are set as $m=25\%$, $b_l=0.1$ Hz, $b_h=0.6$ Hz, and $b_s=0.2$. We set $b_l$ and $b_h$ according to the typical frequency range of human breathing signals. In contrast, $b_s$ is environment-sensitive and we set it empirically.  

To improve robustness, we augment the dataset by shuffling breathing components, ensuring each augmented frame represents a unique radar scene. This increases the dataset to 1,250 recordings, each lasting 30 seconds. Table \ref{tab:dataset} summarizes the dataset.

\begin{table}[!t]
  \centering
  \begin{tabular}{|c|c|c|c|c|c|c|} \hline
    \multirow{2}{*}{\textbf{Class}} & \multicolumn{4}{c|}{\textbf{D1 Set}} & \textbf{D2 Set} & \textbf{D3 Set} \\ \cline{2-7} 
    & \textbf{Train} & \textbf{Valid} & \textbf{Test} & \textbf{Total} & \textbf{Test} & \textbf{Test} \\ \hline 
    Two & 70 & 20 & 40 & 130 & 160 & 50 \\ \hline
    Three & 80 & 20 & 40 & 140 & 150 & 50 \\ \hline
    Five & 70 & 20 & 40 & 130 & 100 & 50 \\ \hline
    Seven & 80 & 20 & 40 & 140 & 100 & 50 \\ \hline
    \textbf{Total} & \textbf{300} & \textbf{80} & \textbf{160} & \textbf{540} & \textbf{510} & \textbf{200} \\ \hline
  \end{tabular}
  \caption{Dataset summary}
  \label{tab:dataset}
  \vspace{-9mm}
\end{table}

\begin{table*}[!t]
\centering
\begin{tabularx}{\textwidth}{|c|c|*{8}{Y|}} \hline
\multirow{2}{*}{\textbf{Dataset}}
&
\multirow{2}{*}{\textbf{Model}}
& 
\multicolumn{3}{c|}{\textbf{Average Accuracy ($\uparrow$)}}
&
\multicolumn{3}{c|}{\textbf{Weighted Average Accuracy ($\uparrow$)}}
&
\multicolumn{2}{c|}{\textbf{Counting Error ($\downarrow$)}}
\\ \cline{3-10}

&
&
\textbf{Precision}
&
\textbf{Recall}
&
\textbf{F1 Score}
&
\textbf{Precision}
&
\textbf{Recall}
&
\textbf{F1 Score}
&
\textbf{MAE}
&
\textbf{MSE}
\\ \hline

\multirow{4}{*}{D1}
&
ViT-only-reg
&
9.1\%
&
25.0\%
&
13.3\%
&
6.2\%
&
17.0\%
&
9.1\%
&
1.2
&
2.1
\\ \cline{2-10}

&
ViT-only-cls
&
64.3\%
&
56.3\%
&
60.0\%
&
67.2\%
&
63.0\%
&
65.0\%
&
0.7
&
\textbf{1.2}
\\ \cline{2-10}

&
\sys-reg
&
36.7\%
&
37.5\%
&
37.1\%
&
26.2\%
&
28.7\%
&
27.4\%
&
1.1
&
2.1
\\ \cline{2-10}

&
\textbf{\sys-cls}
&
\textbf{83.8\%}
&
\textbf{81.3\%}
&
\textbf{82.5\%}
&
\textbf{88.1\%}
&
\textbf{85.6\%}
&
\textbf{86.8\%}
&
\textbf{0.6}
&
2.3
\\ \hline

\multirow{4}{*}{D3}
&
ViT-only-reg
&
\textbf{55.0\%}
&
15.0\%
&
23.6\%
&
58.1\%
&
14.3\%
&
23.0\%
&
1.5
&
3.5
\\ \cline{2-10}

&
ViT-only-cls
&
37.1\%
&
25.0\%
&
29.9\%
&
49.4\%
&
20.2\%
&
28.6\%
&
1.7
&
4.6
\\ \cline{2-10}

&
\sys-reg
&
41.7\%
&
25.0\%
&
31.3\%
&
57.6\%
&
31.7\%
&
40.9\%
&
\textbf{1.0}
&
\textbf{1.6}
\\ \cline{2-10}

&
\textbf{\sys-cls}
&
52.7\%
&
\textbf{45.0\%}
&
\textbf{48.5\%}
&
\textbf{67.6\%}
&
\textbf{53.3\%}
&
\textbf{59.6\%}
&
1.1
&
2.3
\\ \hline

\end{tabularx}
\caption{Overall people counting results of ViT-only and \sys in seen (D1) and unseen (D3) environments}
\label{tab:overall_results}
\vspace{-8mm}
\end{table*}

\subsection{Model Training and Testing}
We train a ViT-B/16 model \cite{conard2023vit} using PyTorch with a $224 \times 224$ input image size and a $16$ patch size. The model is initialized with pre-trained weights from the large-scale ImageNet-21K dataset \cite{ridnik2021imagenetk}. We do not tune the hyperparameters, as they are already optimized for the pre-trained dataset.  

Training is performed using the Stochastic Gradient Descent optimizer and Cross Entropy loss with Soft Targets. Each configuration runs for at least 150 epochs, saving the best model (based on the highest F1 score) every five epochs. Training is accelerated on a single NVIDIA GeForce RTX 2080 Ti GPU (12 GB memory). The model is trained on 70\% of the dataset and tested on the remaining 30\%. During testing, each recording is augmented ten times, and predictions are made using majority voting for final classification.

\begin{table}[!t]
\centering
\begin{tabular}{|c|c|c|c|} \hline
\multirow{2}{*}{\textbf{Class}}
&
\textbf{Density}
&
\multicolumn{2}{c|}{\textbf{F1 Score ($\uparrow$)}}
\\ \cline{3-4}  

&
\textbf{(people per $m^2$)}
&
\textbf{D1 Dataset}
&
\textbf{D3 Dataset}
\\ \hline

2
&
0.7
&
100.0\%
&
80.0\%
\\ \hline

3
&
1
&
75.0\%
&
33.3\%
\\ \hline

5
&
1.7
&
85.7\%
&
44.4\%
\\ \hline

7
&
2.4
&
66.7\%
&
0.0\%
\\ \hline

\textbf{Avg}
&
\textbf{-}
&
\textbf{82.5\%}
&
\textbf{48.5\%}
\\ \hline

\textbf{W. Avg}
&
\textbf{-}
&
\textbf{86.8\%}
&
\textbf{59.6\%}
\\ \hline

Low
&
$<1$
&
87.5\%
&
75.0\%
\\ \hline

High
&
2-3
&
87.5\%
&
83.3\%
\\ \hline

\textbf{Avg}
&
\textbf{-}
&
\textbf{87.5\%}
&
\textbf{82.8\%}
\\ \hline

\textbf{W. Avg}
&
\textbf{-}
&
\textbf{87.5\%}
&
\textbf{80.4\%}
\\ \hline

\end{tabular}
\caption{Class-wise and density-wise F1 score of \sys}
\label{tab:class_results}
\vspace{-9mm}
\end{table}

\subsection{Evaluation Metrics}
We evaluate \sys using average precision, recall, and F1-score across four classes, applying both macro and weighted averaging techniques. The macro average computes the mean of all class-wise results, treating each class equally. \rnote{In contrast, the weighted average assigns different weights to each class, accounting for the greater impact of counting errors in smaller groups (e.g., two people) compared to larger groups (e.g., seven people). To penalize errors in smaller groups more heavily, we assign higher weights to them and lower weights to larger groups, with each class weight set as the reciprocal of its label: $w_2 = 1/2$, $w_3 = 1/3$, $w_5 = 1/5$, and $w_7 = 1/7$.} The weighted precision ($P_w$), recall ($R_w$), and F1-score ($F1_w$) are computed as:
\[
P_w = \frac{\sum_i w_i P_i}{\sum_i w_i},\quad
R_w = \frac{\sum_i w_i R_i}{\sum_i w_i},\quad
F1_w = \frac{2}{1/P+1/R}
\]

\rnote{Additionally, we evaluate the people counting results using Mean Absolute Error (MAE) and Mean Squared Error (MSE).} 

\subsection{Baselines}  
As no static people-counting approaches exist in the literature, we compare our method against a ViT-only baseline introduced in the preliminary study. This baseline model processes the average Range-Azimuth spectrograms of 12 consecutive frames. We implement two versions of the ViT-only model: classification (ViT-only-cls) and regression (ViT-only-reg). Similarly, we develop two versions of \sys: \sys-cls and \sys-reg, using Cross Entropy loss for classification and Mean Squared Error loss for regression.  

For consistency, we also compute precision, recall, and F1-score for regression models. The regression output is rounded to the nearest integer to obtain a predicted people count, which is treated as a class label. Any predicted class not in the true class set is considered an Out-of-Distribution (OOD) class. A test case is classified as a true positive if the predicted regression class matches the true class.

\begin{figure}[!t]
\centering
\subfloat[Seen environment]
{\includegraphics[width=0.2\textwidth]{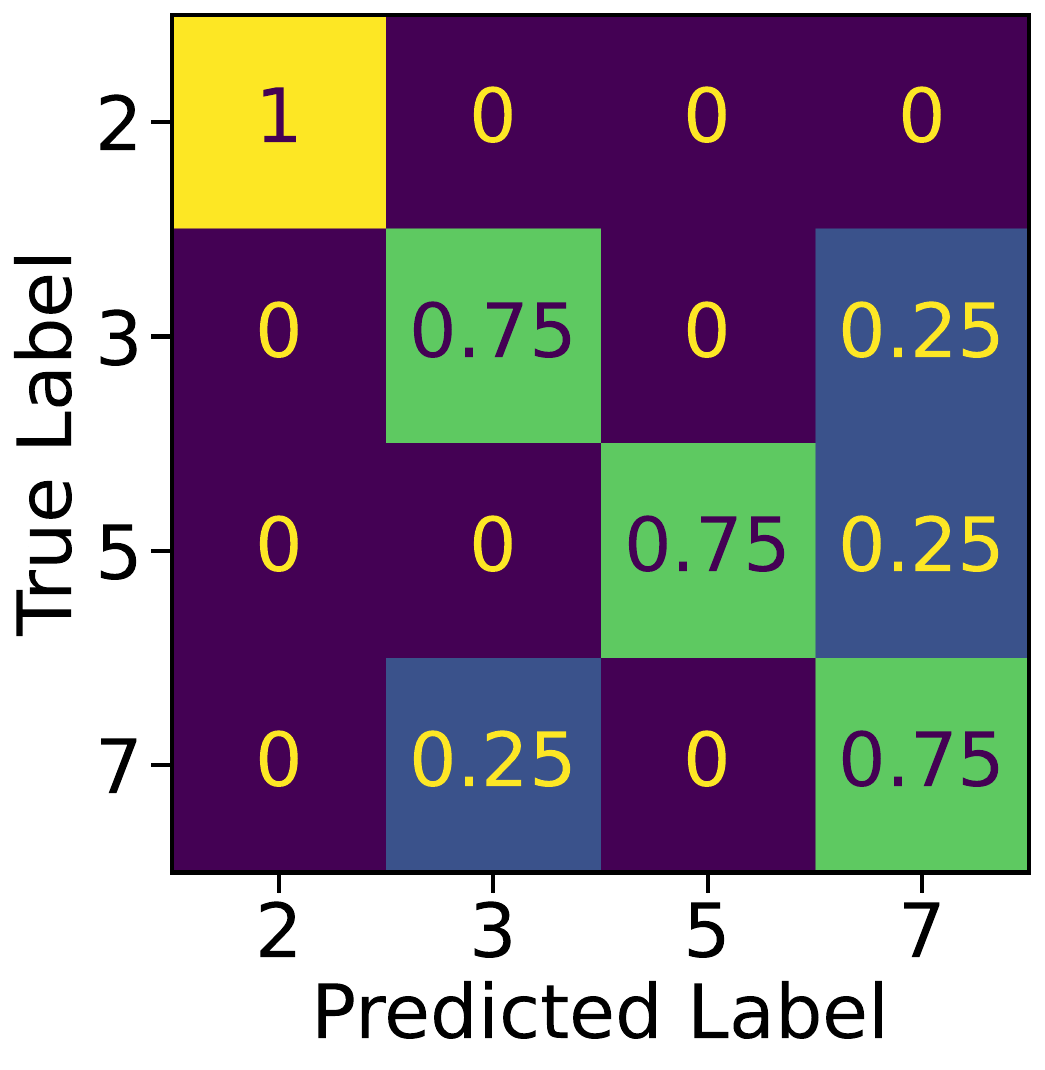}
\label{fig:counting_cm}}
\hfil
\subfloat[Unseen environment]
{\includegraphics[width=0.2\textwidth]{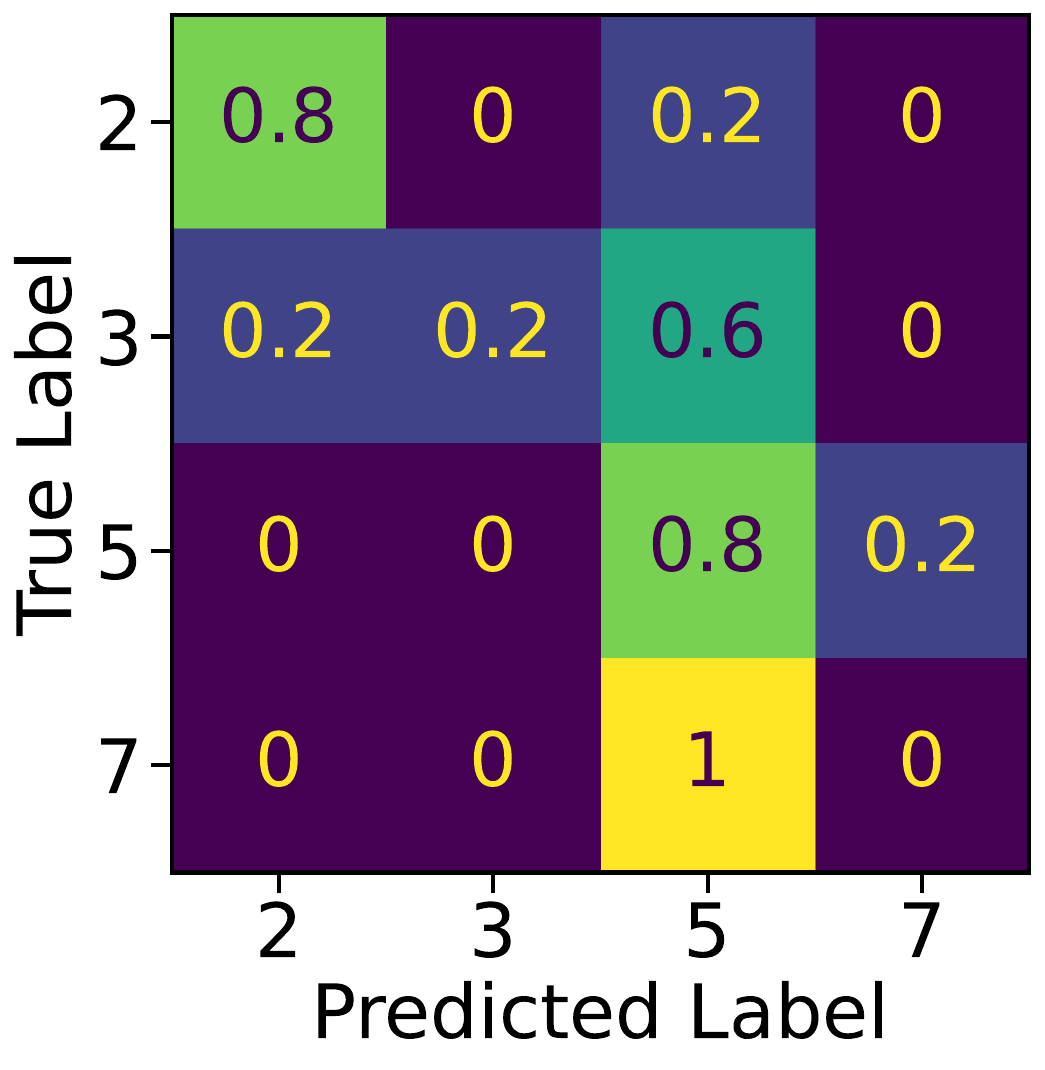}
\label{fig:corridor_cm}}
\vspace{-3mm}
\caption{Confusion matrix of \sys}
\vspace{-3mm}
\end{figure}

\begin{figure*}[!t]
\centering
\subfloat[IPD=1 and Angle=0]
{\includegraphics[width=0.19\textwidth]{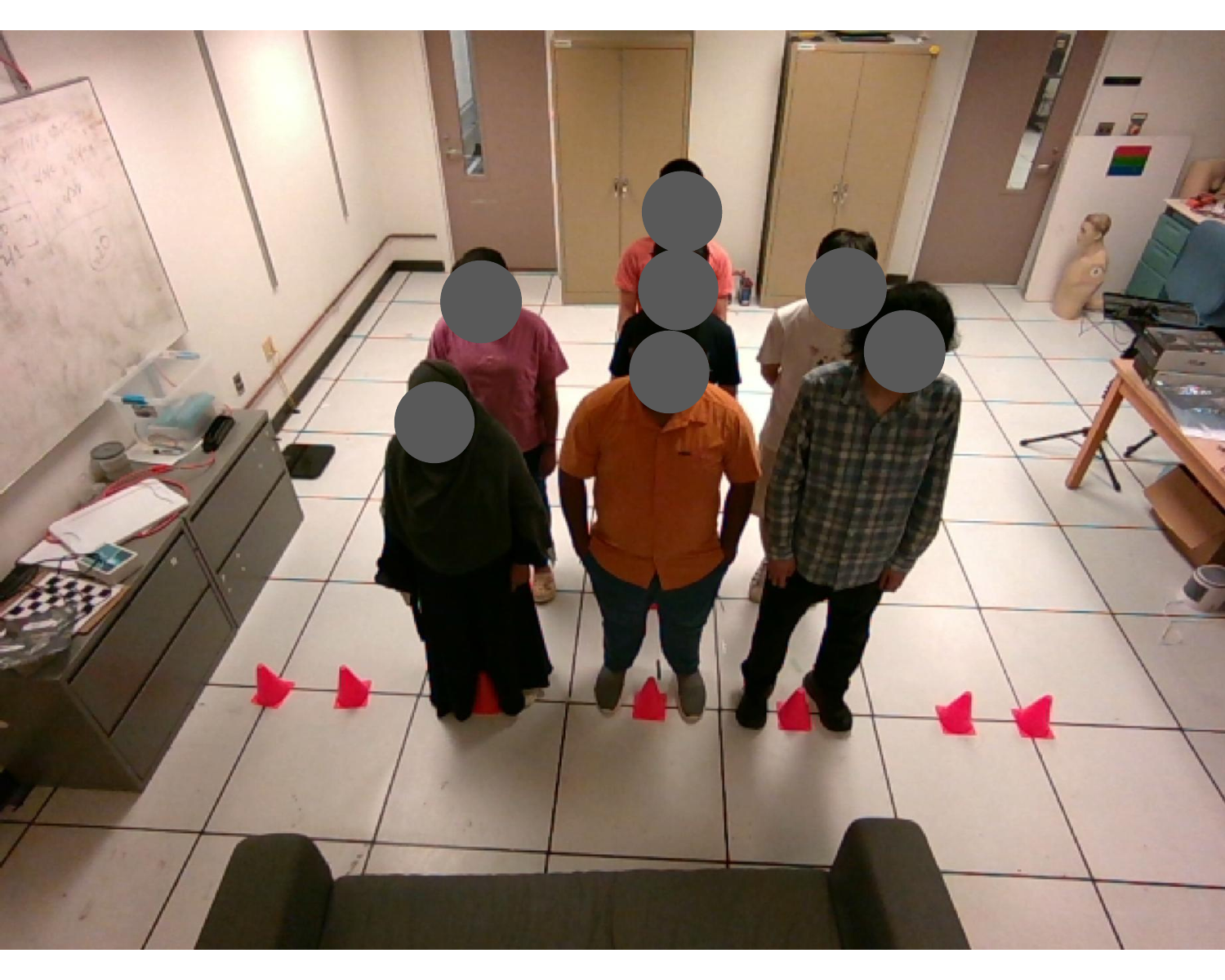}
\label{fig:gap1}}
\hfil
\subfloat[IPD=2 and Angle=0]
{\includegraphics[width=0.19\textwidth]{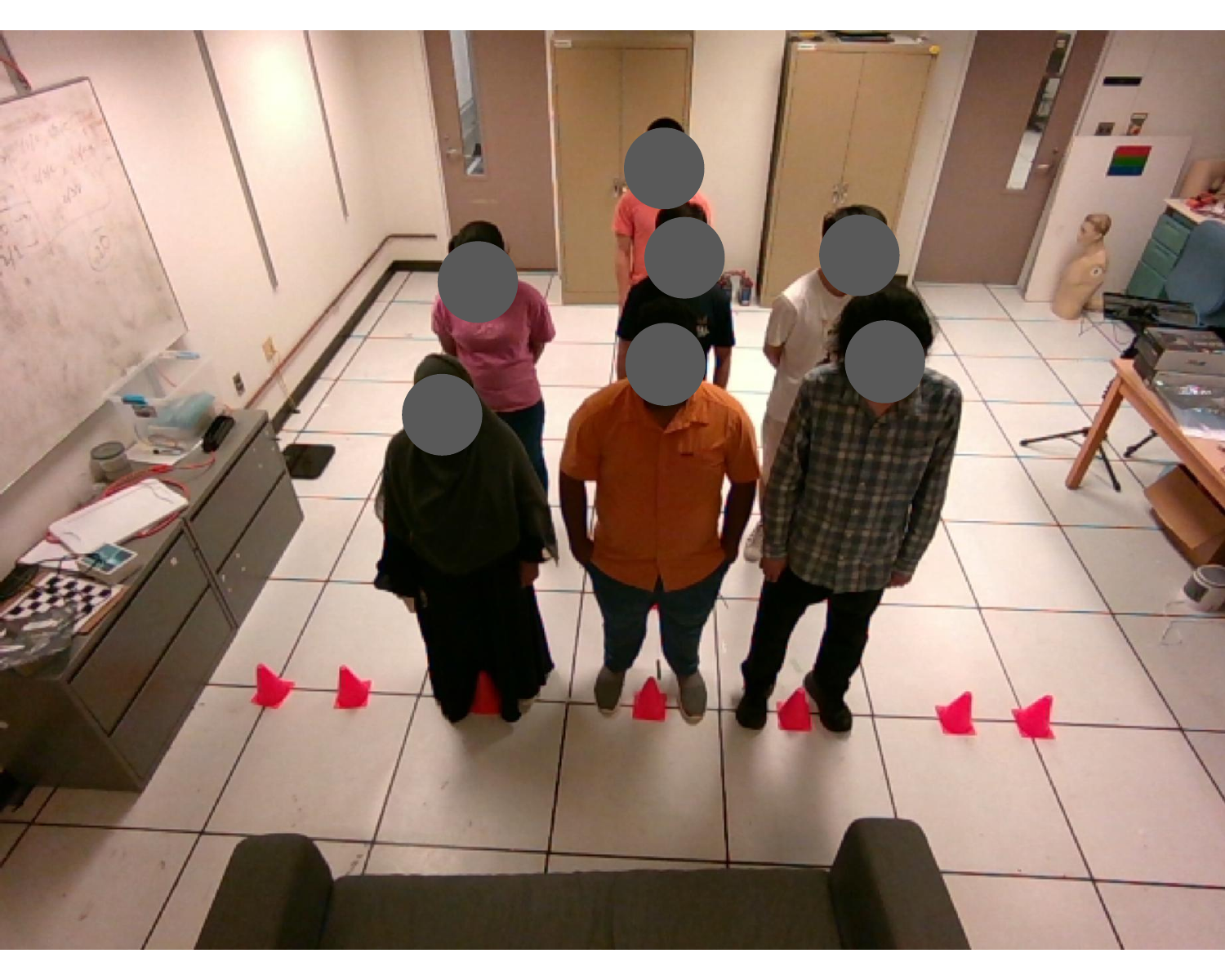}
\label{fig:gap2}}
\hfil
\subfloat[IPD=3 and Angle=0]
{\includegraphics[width=0.19\textwidth]{figures/gap3.pdf}
\label{fig:gap3}}
\hfil
\subfloat[IPD=3 and Angle=-40]
{\includegraphics[width=0.19\textwidth]{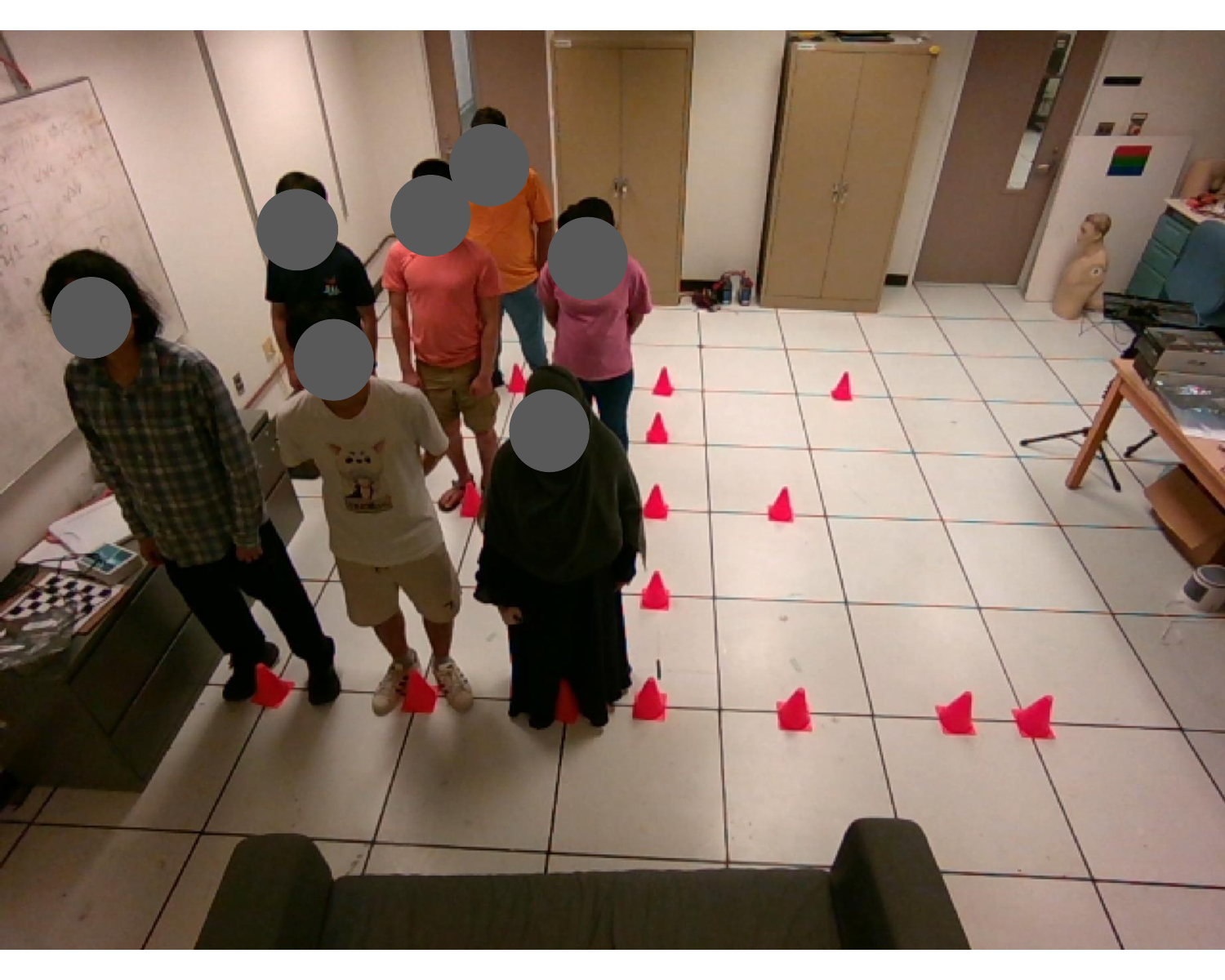}
\label{fig:azi_neg}}
\hfil
\subfloat[IPD=3 and Angle=+40]
{\includegraphics[width=0.19\textwidth]{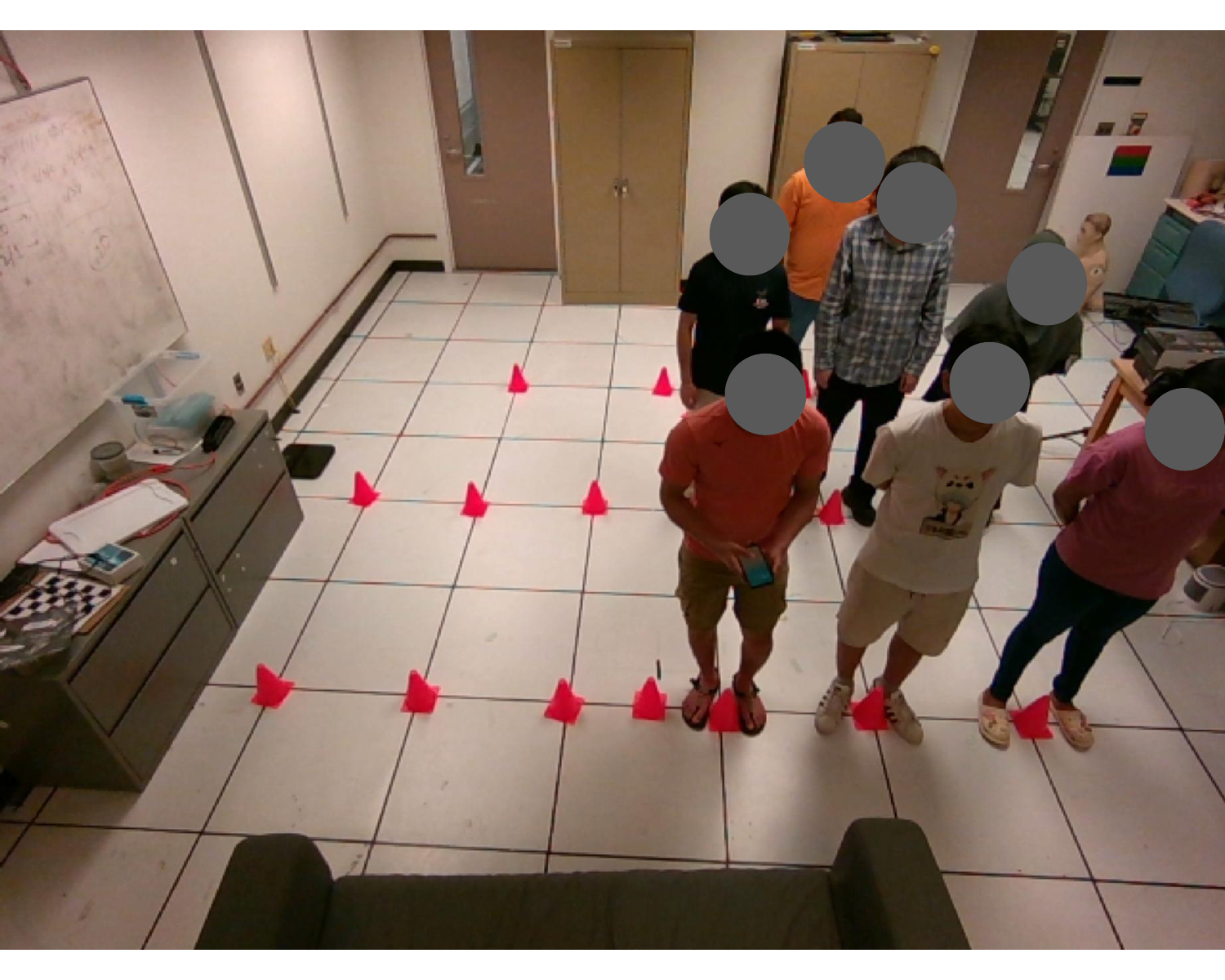}
\label{fig:azi_pos}}
\vspace{-3mm}
\hfil
\subfloat[Impact of inter-personal distance when the angular position is zero degree]
{\includegraphics[width=0.49\textwidth]{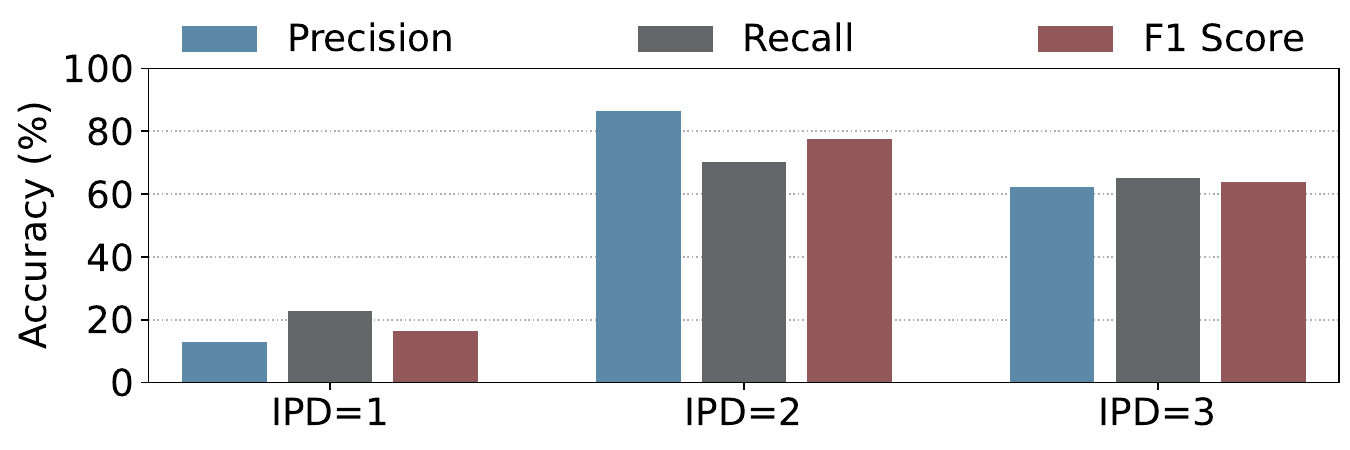}
\label{fig:gap_results}}
\hfil
\subfloat[Impact of angular position when the inter-personal distance is 3 feet]
{\includegraphics[width=0.49\textwidth]{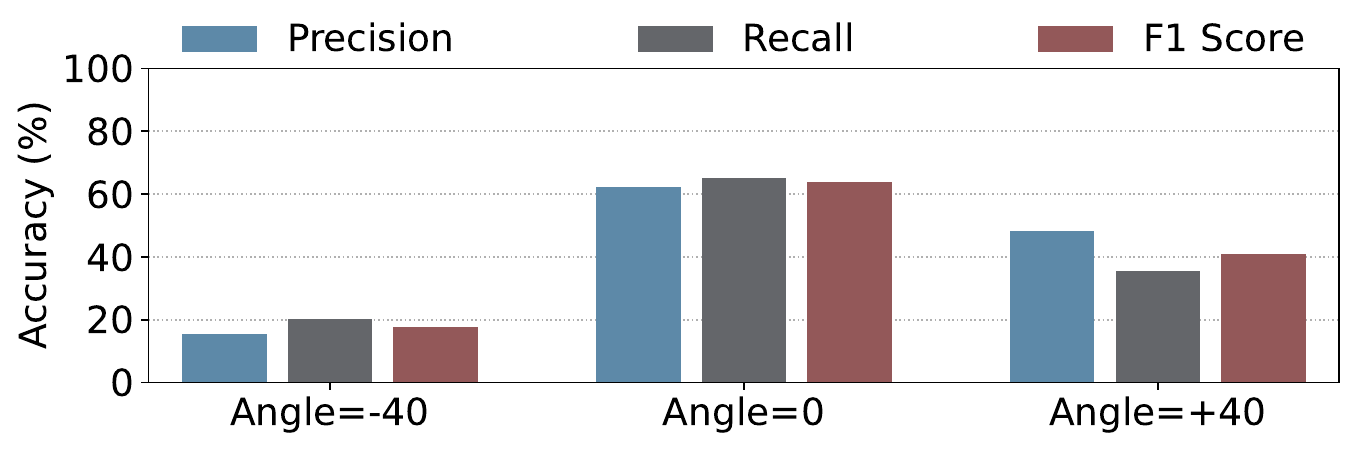}
\label{fig:angle_results}}
\vspace{-3mm}
\caption{Robustness analysis of~\sys in terms of occlusion and angular positions}
\vspace{-3mm}
\end{figure*}

\section{Experimental Results}
\subsection{Accuracy}
We evaluate both ViT-only and \sys models in seen and unseen environments. Since \sys-cls is our final model, we refer to it interchangeably as \sys-cls and \sys throughout the paper.

\parlabel{Accuracy in Seen Environment:}
In this experiment, both the ViT-only and \sys models are trained and tested on the D1 dataset using a 70\%-30\% split, meaning the models are evaluated in a \emph{seen} environment with known users. 

Table \ref{tab:overall_results} presents the weighted classification results. On average, \sys-cls significantly outperforms all other models, achieving 88.1\% precision, 85.6\% recall, and an 86.8\% F1-score. \rnote{Additionally, \sys-cls achieves the lowest MAE among all models. These results confirm the superiority of our classification approach over regression methods as well as its advantage over the end-to-end model.}

\rnote{Table~\ref{tab:class_results} presents the class-wise results, while the corresponding confusion matrix is shown in Figure~\ref{fig:counting_cm}. Most classes are correctly classified in the \emph{seen} environment.} We further analyze performance based on density, categorizing 2-3 people per 3 m$^2$ as low-density and 5-7 people per 3 m$^2$ as high-density. The density-wise classification results in Table~\ref{tab:class_results} indicate that \sys performs well across both density levels. 

The overall results show that the regression model does not outperform the classification model. This is because mmWave-based breathing signals do not scale linearly, overlap significantly in dense environments, and are affected by noise and multipath effects, making accurate regression challenging. Prior studies also favor classification-based approaches for mmWave radar-based people counting \cite{marco2024mmwave}.

\parlabel{Accuracy in Unseen Environment:}
To evaluate generalization, we test \sys in an unseen environment with a new user set, making counting more challenging. Both the ViT-only models and \sys are retrained on all samples from the D1 dataset and evaluated on all samples from the D3 dataset. In this setting, neither the environment nor the user set was previously encountered by the models. Since the new environment alters signal reflection patterns, we adjust the minimum breathing score to $b_s=0.1$ to ensure high-quality breathing signals.

The weighted \emph{unseen} evaluation results are presented in Table~\ref{tab:overall_results}. On average, \sys-cls significantly outperforms all other models with 67.6\% precision, 53.3\% recall, and a 59.6\% F1-score. These results confirm the robustness of our approach over the end-to-end model. Although the regression version of \sys achieves the lowest MAE and MSE, this version is less accurate than the classification version.  

Table \ref{tab:class_results} presents the class-wise results, while the corresponding confusion matrix is shown in Figure \ref{fig:corridor_cm}. Despite operating in an unfamiliar environment, \sys successfully detects most two-person and five-person groups, demonstrating its adaptability to new settings and user sets. The density-based classification results on the D3 dataset, shown in Table \ref{tab:class_results}, further confirm that \sys performs well across different density levels.

\vspace{-1mm}
\subsection{Robustness Analysis}
To assess the robustness of \sys, we evaluate the impact of occlusions, angular positions, motion, body postures, and temporal stability on counting accuracy. Using the same model (\sys-cls) trained for the unseen environment, we test it on the D2 dataset and report the weighted average accuracy. We set the minimum breathing score to $b_s=0.2$. 

\parlabel{Impact of Occlusion:}
We analyze occlusion impact by varying front-to-back inter-personal distances (IPD) from 1 to 3 feet while keeping angular positions constant. Figures~\ref{fig:gap1},~\ref{fig:gap2}, and~\ref{fig:gap3} show a seven-person group at different IPDs, representing worst-case occlusion for \sys. For occluded individuals, we use mmWave radar’s multi-path effects to estimate breathing signals. Figure~\ref{fig:gap_results} shows accuracy improves as IPD increases from 1 to 2 feet due to reduced occlusion but decreases at 3 feet, as greater radar-to-human distance weakens the signal. The effects of occlusion can be mitigated by deploying multiple radars at different locations in a room, which could be explored in future work.

\parlabel{Impact of Angular Position:}
\rnote{We assess the impact of angular positioning (line-of-sight) by varying angles from -40\textdegree~to +40\textdegree~while keeping inter-personal distances constant. Figures \ref{fig:gap3}, \ref{fig:azi_neg}, and \ref{fig:azi_pos} show a seven-person group at different angles, where individuals are not directly facing the radar. As shown in Figure \ref{fig:angle_results}, accuracy is highest at 0\textdegree, where individuals face the radar, maximizing the signal-to-noise ratio (SNR) for better breathing detection. Accuracy decreases as angles increase due to lower SNR. Additionally, left-side interference from nearby furniture causes greater signal distortion, reducing accuracy compared to the right side.}

\begin{figure*}[!t]
\centering
\subfloat[Impact of non-breathing signals]
{\includegraphics[width=0.49\textwidth]{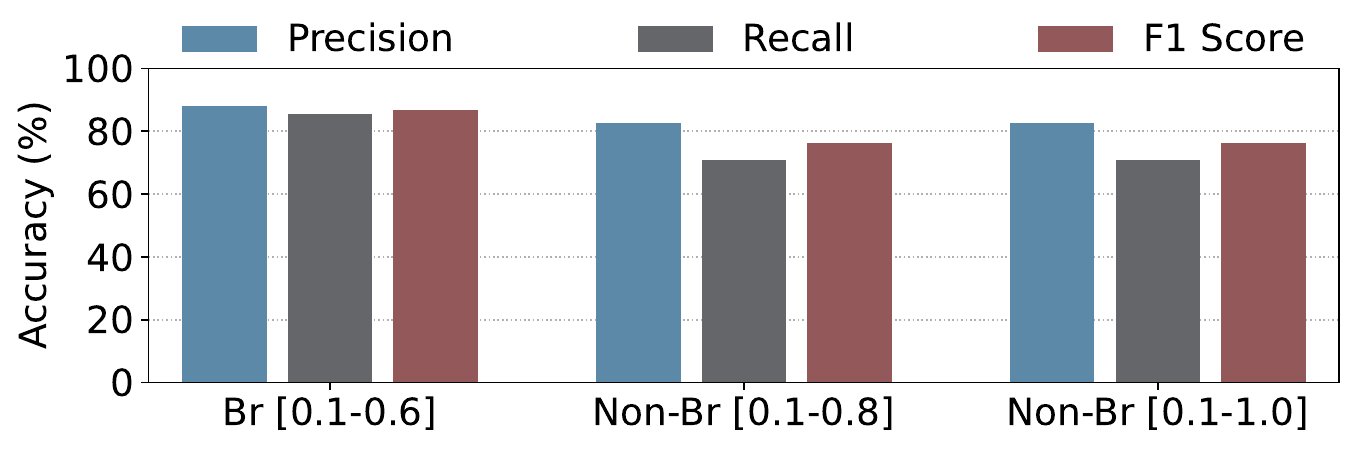}
\label{fig:breathing_band}}
\hfil
\subfloat[Impact of breathing signal quality ($b_s$)]
{\includegraphics[width=0.49\textwidth]{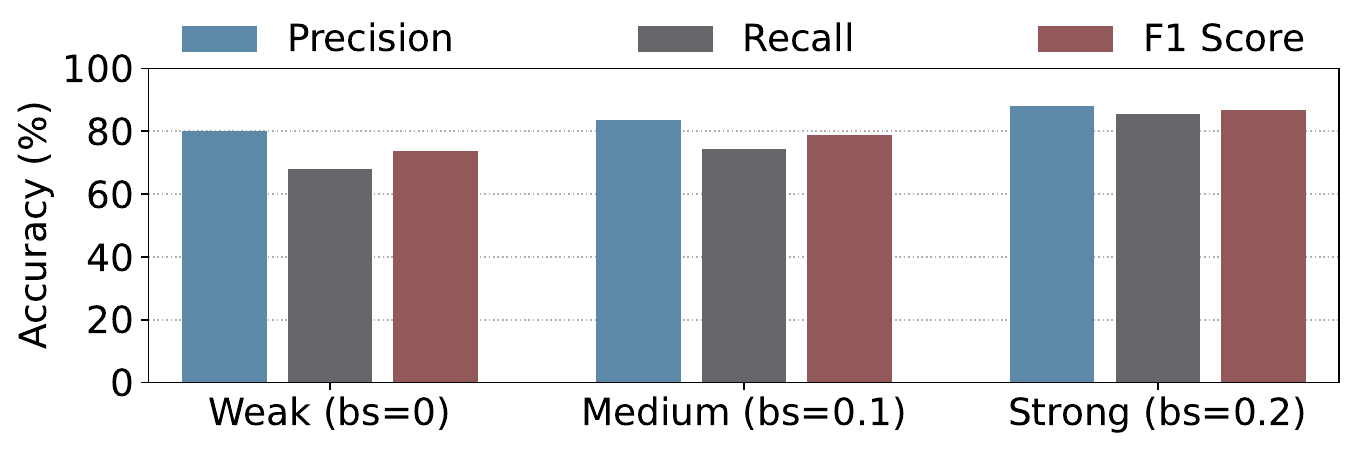}
\label{fig:breathing_score}}
\vspace{-3mm}
\hfil
\subfloat[Impact of frequency selection criteria]
{\includegraphics[width=0.49\textwidth]{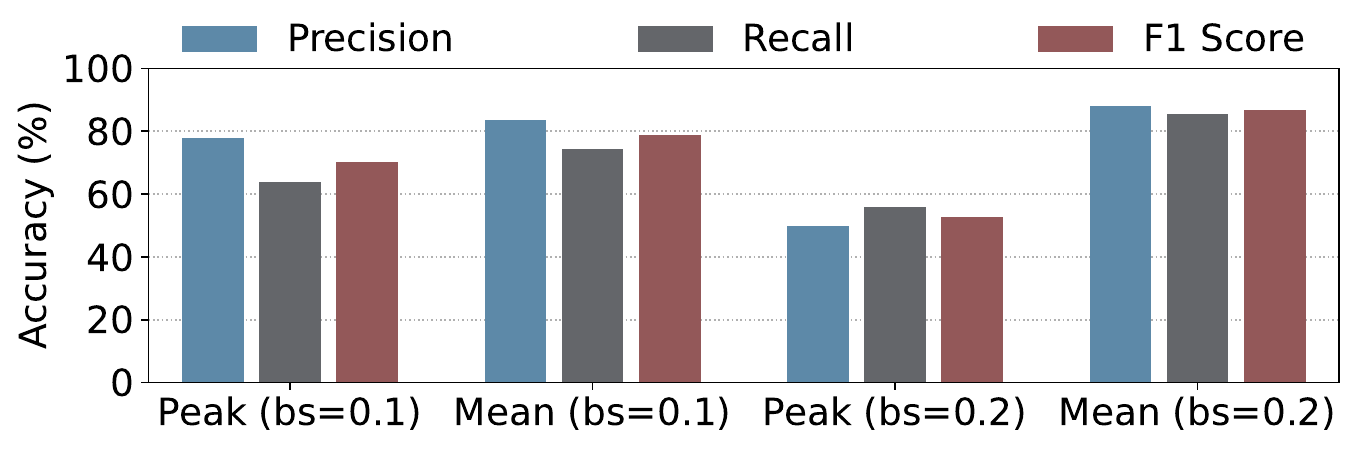}
\label{fig:freq_type}}
\hfil
\subfloat[Impact of data augmentation type]
{\includegraphics[width=0.49\textwidth]{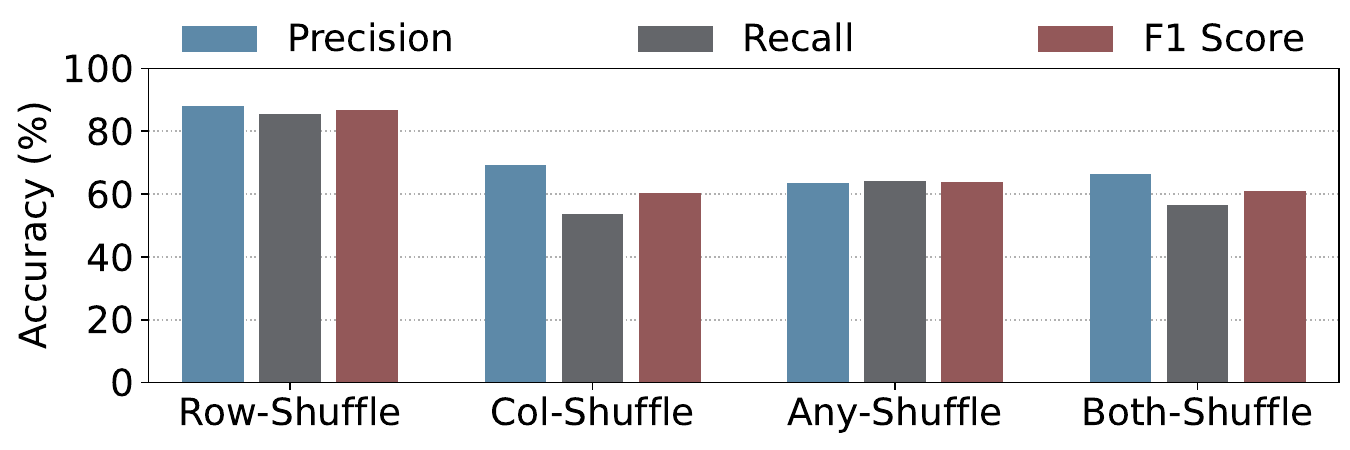}
\label{fig:aug_type}}
\vspace{-3mm}
\caption{Ablation studies of~\sys}
\vspace{-3mm}
\end{figure*}

\parlabel{Impact of Motion:}
\rnote{We assess motion impact through four test scenarios: (i) two participants remain standing, (ii) two stand, then leave the FoV together, (iii) three remain standing, and (iv) three stand, then one leaves the FoV. As dynamic people counting is well-studied \cite{marco2024mmwave}, we discard radar frames with motion and apply \sys to the static frames. Table~\ref{tab:motion_results} shows recall across scenarios, with \sys correctly predicting most cases with 0.4 mean absolute error, demonstrating robustness to partial movement. Accuracy may improve with a more precise motion detection algorithm.}  

\begin{table}[!t]
\centering
\begin{tabular}{|*{5}{c|}} \hline

\multirow{2}{*}{\textbf{Test Scenarios}}
&
\textbf{Total}
&
\textbf{Correct}
&
\multirow{2}{*}{\textbf{MAE}}
&
\multirow{2}{*}{\textbf{MSE}}
\\

&
\textbf{Cases}
&
\textbf{Cases}
&
&
\\ \hline 

two-static
&
4
&
3
&
0.3
&
0.3
\\ \hline 

three-static
&
4
&
3
&
0.5
&
1
\\ \hline 

two-static-two-leave
&
4
&
2
&
0.5
&
0.5
\\ \hline 

three-static-one-leave
&
3
&
2
&
0.4
&
0.4
\\ \hline 

two-sitting
&
2
&
2
&
0
&
0
\\ \hline 

three-sitting
&
2
&
1
&
0.5
&
0.5
\\ \hline 

\end{tabular}
\caption{Impact of motion and body postures}
\label{tab:motion_results}
\vspace{-9mm}
\end{table}

\parlabel{Impact of Body Postures:}
\rnote{We evaluate the impact of body posture on system accuracy by comparing standing and sitting scenarios. Participants were instructed to sit side-by-side in chairs, simulating real-world waiting room conditions. The results for the sitting scenario are presented in Table~\ref{tab:motion_results}. Despite not training on sitting data, \sys accurately predicts most sitting cases with 0.3 mean absolute error, demonstrating its robustness in real-world conditions.}  

\parlabel{Temporal Stability:}
\rnote{We assess the temporal stability of \sys by repeatedly measuring the same group of individuals. Two participants are instructed to stand naturally side by side for approximately 13 minutes while the radar scene is recorded 10 times. \sys accurately predicts the true count in most instances, with a mean absolute error of 0.9. These results demonstrate the robustness of our approach over time.}

\vspace{-3mm}
\subsection{Ablation Studies}
We fine-tune various parameters of \sys to improve breathing signal quality and overall accuracy. Specifically, we assess the impact of non-breathing signals, breathing signal quality, frequency selection criteria, data augmentation, ICA iterations, and antenna resolution on the D1 dataset. 

\parlabel{Impact of Non-Breathing Signals:}
The goal of this experiment is to determine whether non-breathing signals contribute positively to people counting. A signal is classified as non-breathing if its mean frequency, $\Bar{f}$, exceeds 0.6 Hz. To test this, we increase the upper limit of the frequency band from 0.6 to 1.0 Hz in Equation \ref{eq:br}. The results are shown in Figure \ref{fig:breathing_band}, where the accuracy decreases when non-breathing signals are included. This indicates that non-breathing signals lack signature information relevant to identifying individuals.

\parlabel{Impact of Breathing Signal Quality:}
We investigate the impact of breathing signal quality by varying $b_s$ from 0 to 0.2 in Equation \ref{eq:bs}, allowing distorted signals within the breathing frequency band in a controlled manner. The results, shown in Figure \ref{fig:breathing_score}, indicate that including distorted signals does not improve system performance. This suggests that, while these signals fall within the breathing band, they are corrupted by other factors and lose the distinctive information needed to identify individuals.

\parlabel{Impact of Frequency Selection Criteria:}
An argument could be made for using \emph{peak} frequency instead of \emph{mean} frequency for breathing signal extraction. To investigate this, we evaluate both criteria. The results, shown in Figure \ref{fig:freq_type}, indicate that mean frequency provides better accuracy than peak frequency. This is likely because the source signals lack a distinct, singular frequency.

\begin{figure}[!t]
\centering
\includegraphics[width=0.3\textwidth]
{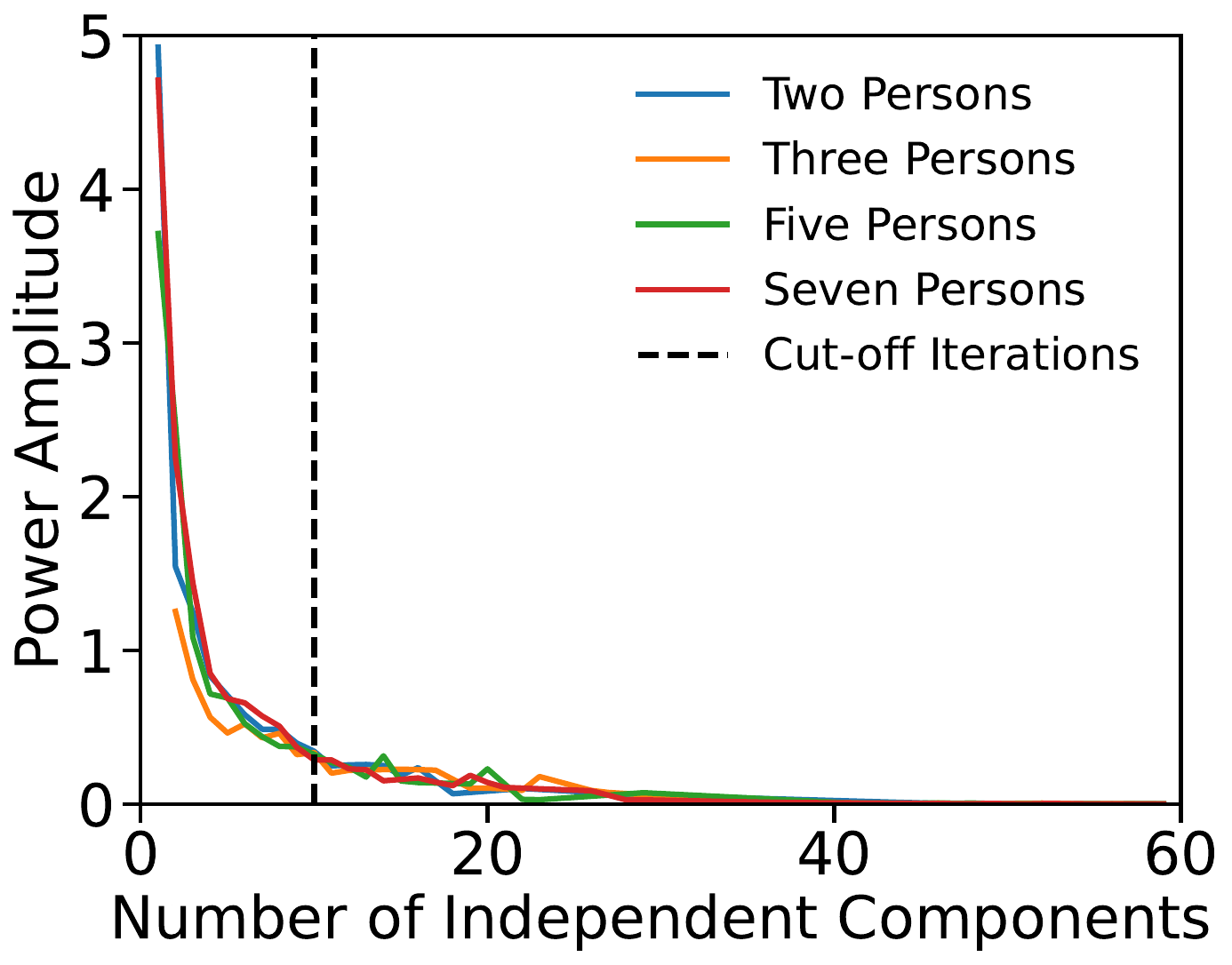}
\vspace{-3mm}
\caption{Impact of ICA iterations}
\label{fig:ic_cost}
\vspace{-3mm}
\end{figure}

\parlabel{Impact of Data Augmentation:}
We augment the breathing profiles by shuffling the rows to expand the dataset. However, we could also shuffle the columns, or both rows and columns. To assess the impact, we apply four augmentation types, with the results shown in Figure \ref{fig:aug_type}. Row shuffling yields the highest accuracy, as the order of breathing components does not affect the radar scene. In contrast, column shuffling disrupts spatial consistency, leading to a drop in accuracy.

\parlabel{Impact of ICA Iterations:}
\rnote{We examine the impact of ICA iterations on breathing component separation. In each iteration, valid source signals are extracted using Equations~\ref{eq:br} and~\ref{eq:bs}, and analyzed in the frequency domain by averaging peak power amplitudes. Figure~\ref{fig:ic_cost} shows that increasing iterations reduces peak power amplitudes, attenuating valid signals. Additionally, ICA's computational complexity grows with the number of independent components. To balance signal quality and efficiency, ICA iterations must be limited. Effective separation requires at least as many iterations as the number of breathing sources, equal to the individuals in the radar scene. Based on this trade-off, we set $n=10$.}

\parlabel{Impact of Antenna Resolution:}
\rnote{We evaluate antenna resolution impact by simulating the xWR1843 device (3TX/4RX). Disabling three radar chips reduces virtual antennas from 192 to 12. Under this low resolution, \sys achieves 50.6\% precision, 73.9\% recall, 60.1\% F1 score, and 1.3 mean absolute error. With fewer azimuth dimensions, radar points merge, hindering fine-grained micro-motion extraction and degrading performance. This experiment highlights the need for high-resolution hardware in static and dense people counting with mmWave radar.}

\section{Discussion}


\parlabel{Environmental Sensitivity:}
We collected our dataset in two distinct indoor environments: a lab and a corridor. The lab environment resembles a private office with closed doors and furniture, while the corridor simulates a public space with an open layout. Our findings indicate that \sys's performance varies across environments due to differences in signal reflection patterns, which affect multipath propagation. To improve real-world generalization, techniques such as domain adaptation \cite{zhang2021domain}, data augmentation \cite{yan2023mmgesture}, and few-shot learning \cite{wang2023rf} could be applied; however, we leave their exploration for future work.

\parlabel{Out-of-Distribution Cases:}
\sys's current implementation is trained to count up to seven individuals. In scenarios where more than seven people are present---an unusually crowded setting with little to no personal space within the three-square-meter area---the current implementation defaults to the highest number of individuals it has been trained for, which in this case is seven. To address out-of-distribution cases, prior work has explored reconstruction-based methods \cite{kahya2024hood}, distance measurement \cite{choi2022fmcw}, and metric learning \cite{stadelmayer2023out}, which can be integrated to enhance \sys’s generalization. 
\section{Related Work}

\parlabel{Camera Sensor:} 
Many people-counting approaches leverage deep learning to count individuals in camera images. For example, \cite{lin2022boosting} proposed a \emph{Multifaceted Attention Network} (MAN) to enhance transformer models for crowd counting, while \cite{chen2022counting} combined CNN and transformer models to handle varying densities. Additionally, \cite{shu2022crowd} introduced a frequency-domain approach, distinct from traditional spatial-domain methods. Although these camera-based techniques achieve high accuracy, they rely on visual data, raising privacy concerns and performing poorly in low-light or adverse weather conditions, such as fog or rain. This has led to the development of privacy-preserving sensor-based crowd-counting systems.  

\parlabel{Thermal Sensor:} 
\rnote{Thermal imaging techniques \cite{xie2023efficient, hagenaars2020single} detect heat signatures, offering a privacy-preserving alternative. However, they still pose privacy concerns, as they can capture health-related information \cite{sayed2022deep}. Their accuracy is also affected by environmental factors such as room temperature \cite{bao2021cnn} and lighting \cite{liu2021cross}, making them unreliable in temperature-controlled spaces. Consequently, they are primarily used for intrusion detection in low-light environments \cite{thermal2025blog}. Additionally, in dense settings, overlapping heat signatures complicate individual distinction, posing a significant challenge for thermal-based people counting \cite{collini2024flexible, pan2023cginet}.}

\parlabel{Doorway Sensor:} 
\rnote{Doorway sensors offer a privacy-preserving alternative for people counting \cite{yun2023gan, zhang2022development}. They track individuals entering and exiting a space and are commonly installed at entrances of public places such as shopping malls, supermarkets, and museums. However, their primary limitation is that they must be placed at entry points \cite{korany2021counting}. These sensors provide only an aggregate count of individuals who entered through the door but cannot determine how many people are present in specific subspaces or zones within a room. Additionally, most doorway sensors operate within a fixed beam coverage distance, making them unsuitable for entrances and doorways that do not meet specific dimensional requirements \cite{door2025blog}.}

\parlabel{mmWave Radar:} 
mmWave radar offers a privacy-preserving solution with high-resolution sensing, even in no-light conditions. However, its application for people counting remains underexplored. \cite{marco2024mmwave} proposed a point-cloud-based approach for outdoor people counting using DBSCAN and PointNet \cite{qi2017pointnet}, but it is limited to detecting moving individuals. Similarly, \cite{ren2023grouped} introduced a grouped counting method using cadence velocity diagrams and statistical classifiers, yet it assumes individuals are walking. \cite{weiss2020improved} explored a vital-sign-based approach utilizing multi-view micro-Doppler signatures, but it excludes dense settings such as side-by-side standing. Moreover, it requires three radars and is limited to counting up to four individuals. Next, we discuss additional radar-based vital signal estimation methods.
 
\parlabel{Vital Signal Estimation:} 
Vital signals such as breathing and heartbeats can be detected wirelessly using mmWave radar by capturing tiny chest displacements. Several techniques have been proposed for this purpose. \cite{wang2020remote} and \cite{gao2022real} introduced a phase-difference technique to extract vital signs from radar signals for a single individual positioned near the radar, but their accuracy degrades with slight movement. To overcome this limitation, \cite{zhang2023pi} proposed a physiology-inspired technique that accounts for micro-level random body movements while detecting vital signs for one person. Additionally, \cite{wang2020vimo} developed a vital signal estimation method for multiple individuals, though it requires them to remain in specific positions. Consequently, these approaches are unsuitable for robust people counting, where individuals may stand freely within the radar’s field of view (FoV).  
\section{Conclusion}  
Existing radar-based people-counting methods ensure privacy but rely on motion for detection. In this paper, we demonstrate that accurate counting is possible without noticeable movement by leveraging mmWave radar to capture and analyze micro-motion signals, particularly breathing. Our system, \sys, extracts spatial breathing patterns and employs a foundation model to estimate the number of individuals in the radar scene. Evaluations across two environments show that \sys achieves an 87\% average F1 score and 0.6 mean absolute error in a known environment, and a 60\% F1 score and 1.1 mean absolute error in an unseen environment, demonstrating its robustness. \sys accurately counts up to seven individuals in a three-square-meter area, even with zero side-by-side spacing and only one meter of front-to-back distance. Future work will focus on expanding dataset diversity across different environments and user groups to further enhance accuracy, robustness, and generalizability.

\begin{acks}
This work was partially supported by NSF CAREER Award \#2047461.
\end{acks}

\balance
\bibliographystyle{abbrv}
\bibliography{bib}

\end{document}